\documentclass[sort,compress,square,comma]{article}

\PassOptionsToPackage{numbers,sort&compress,square,comma}{natbib}
\usepackage[main, final]{neurips_2026}

\usepackage[utf8]{inputenc}
\usepackage[T1]{fontenc}
\usepackage{hyperref}
\usepackage{url}
\usepackage{booktabs}
\usepackage{amsfonts}
\usepackage{nicefrac}
\usepackage{silence}
\usepackage{microtype}
\usepackage{xcolor}

\usepackage{amsmath}
\usepackage{amssymb}
\usepackage{mathtools}
\usepackage{amsthm}
\usepackage{xspace}
\usepackage{multirow}
\usepackage{comment}
\usepackage{makecell}
\usepackage{subcaption}

\usepackage{url}
\usepackage{enumitem}
\usepackage{wrapfig}

\usepackage[capitalize,noabbrev]{cleveref}

\theoremstyle{plain}

\theoremstyle{definition}

\theoremstyle{remark}

\def\eg{\emph{e.g.,}\xspace}

\def\vs{\emph{vs.}\xspace}

\newcommand{\VideoLLMs}{\text{VideoLLMs}\xspace}
\newcommand{\VideoLLM}{\text{VideoLLM}\xspace}

\newcommand{\VideoSTF}{\textsc{VideoSTF}\xspace}

\title{\VideoSTF: Stress-Testing Output Repetition\\ in Video Large Language Models}

\author{
Yuxin Cao\textsuperscript{1} \quad
Wei Song\textsuperscript{2}\thanks{Corresponding authors: Wei Song (\href{mailto:w.song@griffith.edu.au}{w.song@griffith.edu.au}) and Jin Song Dong (\href{mailto:dcsdjs@nus.edu.sg}{dcsdjs@nus.edu.sg}).} \quad
Shangzhi Xu\textsuperscript{3} \quad
Jingling Xue\textsuperscript{3} \quad
Jin Song Dong\textsuperscript{1}\footnotemark[1]
\\
\\
\textsuperscript{1}\textit{National University of Singapore, Singapore} \\
\textsuperscript{2}\textit{Griffith University, Australia} \\
\textsuperscript{3}\textit{University of New South Wales, Australia}
}

\begin{document}

\maketitle

\begin{abstract}
Video Large Language Models (\VideoLLMs) have achieved strong performance on video understanding tasks, yet existing benchmarks evaluate only \emph{what} models predict, leaving the stability of \emph{how} they generate largely unexamined. We surface a previously underexplored generation failure of \VideoLLMs{}, defined as \textit{output repetition}, in which the decoder collapses into self-reinforcing loops of repeated phrases or sentences, and present \VideoSTF, a benchmarking framework for systematically measuring, stress-testing, and exploiting this failure mode. 
\VideoSTF formalizes repetition with three complementary $n$-gram-based metrics, ships a standardized testbed of 10,000 diverse videos, and provides a library of controlled temporal stressors. 
Across 10 advanced \VideoLLMs, \VideoSTF reveals four key findings: (i) repetition is pervasive on unperturbed videos and stable across commonly used frame counts, with repetition rates up to 91\%;
(ii) it spans a severity spectrum from mild redundancy to token-cap loops, and is highly amplified by temporal perturbations;
(iii) temporal stressors form a practical black-box attack surface, flipping benign videos into repetitive ones with tens of queries and high attack success rates (up to 98\%),
and (iv) repetition is not explained by visual redundancy, its amplification tracks local temporal disruption, and only repetition penalties reduce it among common mitigations such as top-$k$ sampling, input filtering, and prompt variation, but increasing the penalty weakens visual grounding.
\VideoSTF reframes generation stability as a useful and complementary evaluation axis for \VideoLLMs and provides the tools to study it. 
The project page is available at \url{https://videostf.github.io/}. 

\end{abstract}

\section{Introduction}

Video has become a pervasive modality in the modern digital era, and the ability of artificial intelligence systems to understand videos has reached a new peak with the emergence of Video Large Language Models (\VideoLLMs)~\cite{wang2024qwen2,chen2024InternVL2.5,wang2025internvl3}. Benefiting from their strong temporal reasoning capability and powerful multimodal representation learning, \VideoLLMs have been widely adopted in video understanding tasks such as video captioning~\cite{yang2023vid2seq,chen2024sharegpt4video} and question answering~\cite{li2023videochat,tang2025aks}. 

\begin{figure}[t]
    \centering
    \includegraphics[width=0.98\linewidth]{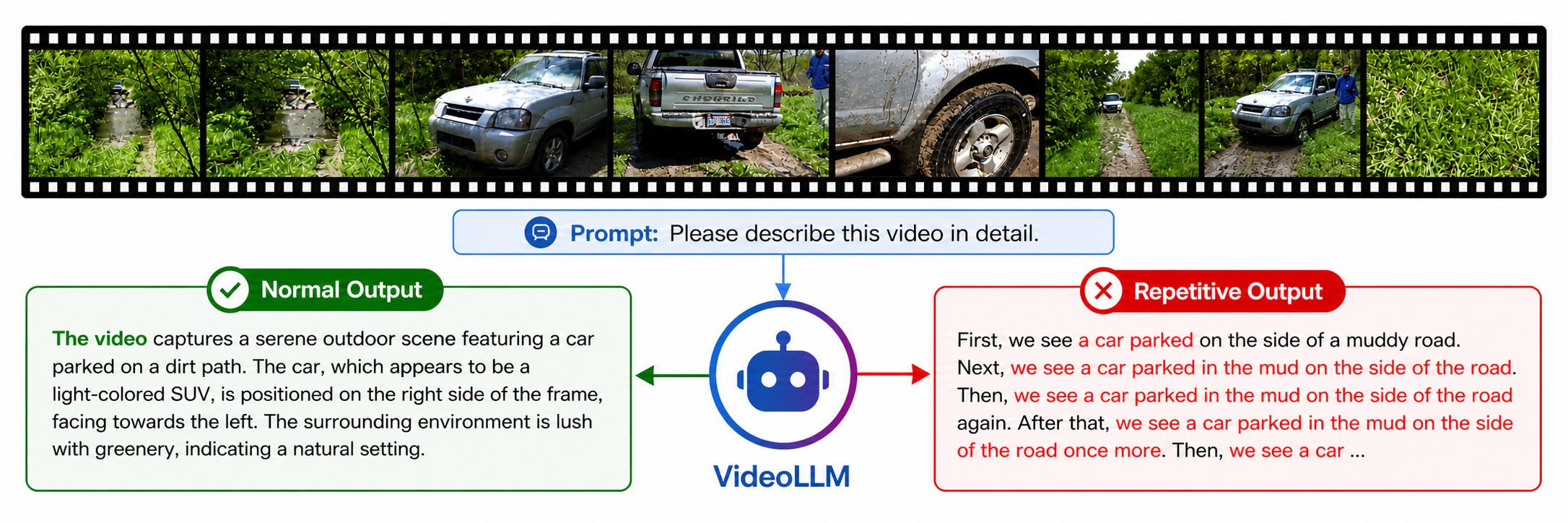}
    \vspace{-3mm}
    \caption{Examples of normal and repetitive outputs.}
    \label{fig:demo}
    \vspace{-5mm}
\end{figure}

However, we identify \textbf{output repetition}, a severe autoregressive generation failure in \VideoLLMs arising from interactions between temporal visual inputs and language generation. 
As shown in Figure~\ref{fig:demo}, a typical normal output produces text that accurately and coherently describes a video, whereas a repetitive output degenerates into repeated phrases or sentences, which can degrade response quality and user experience. In extreme cases, such repetition persists until the model reaches its output token limit, resulting in unnecessary computation and enabling adversarial exploitation~\cite{gao2024inducing}, such as denial-of-service, undermining the reliability of \VideoLLMs.
While numerous benchmarks have been conducted for \VideoLLMs~\cite{ning2025video,li2024mvbench,li2025benchmark,wu2024longvideobench,cao2025video,cao2026failures,li2025vidhalluc,zhou2025flattery}, they primarily focus on task correctness, factual errors, and hallucinations, overlooking generation repetition.

In this work, we systematically study this generation failure in \VideoLLMs and introduce \VideoSTF, a benchmark for measuring, probing, and exploiting output repetition.
\VideoSTF first formalizes output repetition using three $n$-gram-based metrics: a repetition rate that captures whether repetition occurs, a repetition intensity that quantifies the extent of duplicated patterns, and an entropy-based metric that quantifies repetition through lexical diversity.
Furthermore, \VideoSTF provides a standardized video testbed of 10,000 examples, with a temporal stressor library that defines a set of temporal transformations for stress-testing output repetition in \VideoLLMs.

Using \VideoSTF, we conduct three tests: pervasive testing to characterize repetition under natural inputs, temporal stress testing to probe sensitivity to temporal transformations, and adversarial exploitation to assess whether repetition can be actively induced as a black-box attack. We evaluate 10 \VideoLLMs, 
including seven LLM-centric models, LLaVA-Video-7B-Qwen2, LLaVA-Video-7B-Qwen2-Video-Only~\cite{zhang2024LLaVA-Video}, LLaVA-NeXT-Video-7B, LLaVA-NeXT-Video-7B-DPO, LLaVA-NeXT-Video-32B-Qwen~\cite{zhang2024llavanextvideo}, VideoLLaMA2~\cite{cheng2024videollama2}, and ShareGPT4Video~\cite{chen2024sharegpt4video}, as well as three native multimodal models, InternVL3.5-8B~\cite{wang2025internvl3}, Qwen3-VL-8B-Instruct~\cite{bai2025qwen3vl}, and Molmo2-8B~\cite{clark2026molmo2}.

\textbf{Pervasive Testing.} Using the standardized testbed, \VideoSTF benchmarks each \VideoLLM under varying numbers of sampled frames, treating temporal granularity as a controlled variable. Across all 10 models, repetition on unperturbed videos is widespread and stays stable across commonly used frame counts, indicating a pervasive failure mode that persists under the frame settings typical of inference. The repetition typically involves recurring references to the same people, objects, or scenes. Moreover, characteristic looping expressions such as ``continues to'' are prevalent in the video captions used to train some of these models, suggesting a model-internal tendency to enter self-reinforcing repetition patterns.

\textbf{Temporal Stress Testing.} If repetition stems from how \VideoLLMs aggregate temporally similar visual cues, perturbing temporal structure while preserving semantics should sharpen the effect. We therefore apply five stressor families (Add, Delete, Replace, Reverse, and Shuffle) and compare each model's outputs to its original one. Repetition rates rise under most stressors, reaching over 90\% in the most affected cases, indicating a substantial sensitivity to structured temporal perturbations.

\textbf{Adversarial Exploitation.} If a single temporal stressor amplifies repetition population-wide, the same stressor may also \emph{flip} individual benign videos into repetitive ones. We therefore cast the stressor library as a query-efficient black-box attack: under a realistic threat model in which an attacker can modify the sampled frames used for inference but has no access to model internals, attacks succeed with tens of queries (attack success rate up to 98\%). Output repetition is therefore not merely a diagnostic artifact \emph{measured} by \VideoSTF but an exploitable failure mode of modern \VideoLLMs.

Beyond the three benchmarking tests, \VideoSTF further probes the \emph{drivers} of repetition and uncovers four key findings: (i) repetition spans a severity \emph{spectrum} and lowers the information density of descriptions; (ii) it is not explained by visual redundancy, since neither highly redundant inputs nor redundancy filtering raise or remove it; (iii) \emph{local} temporal disruption is the feature-level signal that tracks its amplification; and (iv) a training-data audit links repetition to video-caption supervision, while among common mitigation strategies, only increasing the repetition penalty reduces repetition, albeit at the cost of weaker visual grounding. These findings extend \VideoSTF from a measurement tool to a diagnostic framework that points toward components that future work needs to revisit.

In summary, we make the following contributions:
\begin{itemize}[leftmargin=*, topsep=0pt, itemsep=0pt]
\item We identify output repetition as an underexplored generation-stability failure in \VideoLLMs and introduce \VideoSTF, a benchmarking framework designed to expose and measure this failure through three complementary $n$-gram-based metrics, a standardized testbed of 10,000 diverse videos, and a library of controlled temporal stressors.

\item Using \VideoSTF, we benchmark 10 representative \VideoLLMs and show that repetition is pervasive on unperturbed videos (repetition rate up to 91\%), stable across commonly used frame counts, and strongly amplified by temporal perturbations even when semantic content is preserved.

\item We demonstrate that the same temporal stressors constitute a practical black-box attack surface, flipping benign videos into repetitive ones with tens of queries (attack success rate up to 98\%), elevating output repetition from a quality issue to a reliability and security concern.

\item Beyond measurement, controlled analyses show that repetition is not explained by visual redundancy, and that $\Delta$repetition correlates with $\Delta$adjacent-frame similarity alone. We further trace the looping style to video-caption training data and quantify the grounding cost of repetition penalties.
\end{itemize}

\section{Related Work}

\textbf{\VideoLLMs.}
\VideoLLMs~\cite{zhang2024LLaVA-Video, zhang2024llavanextvideo, cheng2024videollama2, chen2024sharegpt4video, wang2025internvl3} extend LLMs to video understanding by conditioning language generation on temporally ordered visual inputs. They support tasks such as video captioning, video question answering, and video summarization. Existing \VideoLLMs broadly fall into two paradigms. LLM-centric models retain a pretrained LLM as the generator and incorporate video through visual encoders and temporal projection modules~\cite{zhang2024llava-hound, zhang2024llavanextvideo, zhang2024LLaVA-Video, lin2024video-llava, li2024llama-vid, cheng2024videollama2, zhang2025videollama3}. In contrast, native multimodal models such as InternVL3.5-8B~\cite{wang2025internvl3} and Qwen3-VL-8B-Instruct~\cite{bai2025qwen3vl} adopt unified multimodal backbones for jointly modeling vision and language. Despite architectural differences, both paradigms implicitly assume stable autoregressive generation once visual understanding succeeds.

\textbf{VideoLLM Benchmarks.}
Existing \VideoLLMs benchmarks evaluate video understanding accuracy~\cite{ning2025video, li2024mvbench, li2025benchmark}, temporal reasoning~\cite{wu2024longvideobench}, and semantic reliability, including factual errors~\cite{cao2025video}, safety~\cite{cao2026failures}, hallucinations~\cite{li2025vidhalluc}, and behavioral biases such as sycophancy~\cite{zhou2025flattery}. These benchmarks primarily assess \emph{what} a model predicts, while largely overlooking \emph{how} generation unfolds. In particular, the stability of autoregressive generation is rarely examined, leaving output repetition—a severe degeneration phenomenon in \VideoLLMs{}—largely unmeasured and unaccounted.

\textbf{Output Repetition.}
Output repetition has been increasingly studied in text-only LLMs, where it has been attributed to intrinsic model distributions~\cite{fu2021theoretical}, repetitive patterns in training data~\cite{li2023repetition}, and identifiable internal activation features~\cite{yao2025understanding}. These studies establish repetition as a fundamental challenge in autoregressive generation that cannot be fully resolved by decoding strategies alone, even under deterministic decoding. However, they do not consider video-conditioned generation, where repetition can arise from interactions between temporal visual inputs and language decoding. In contrast to prior work, we study output repetition as a video-conditioned generation failure and introduce a dedicated framework for measuring and stress-testing this phenomenon in \VideoLLMs.

\begin{figure*}[t]
    \centering
    \includegraphics[width=0.92\linewidth]{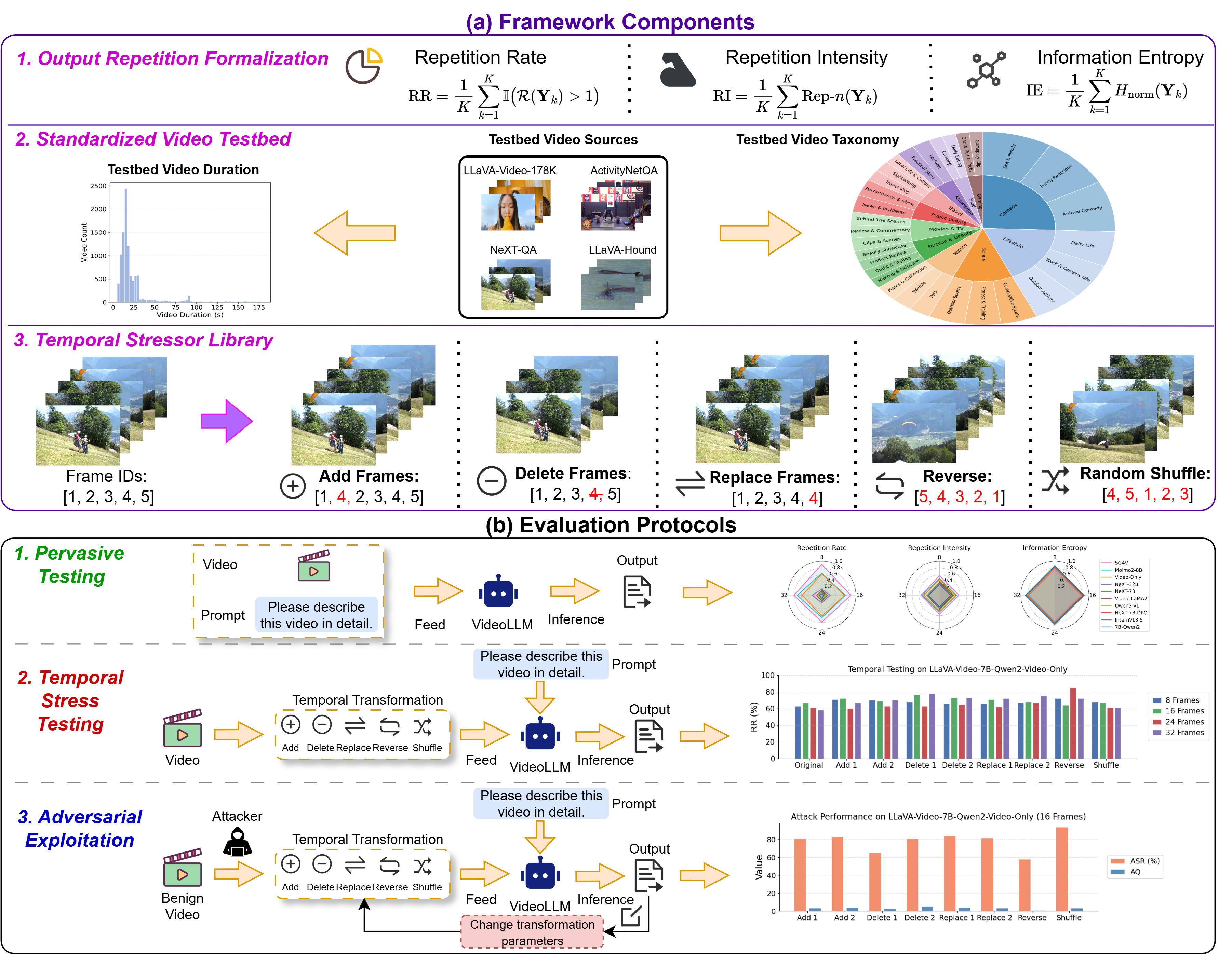}
    \caption{\VideoSTF. (a) \textit{Framework Components}. \VideoSTF comprises three $n$-gram-based repetition metrics, a standardized testbed of 10,000 videos with diverse durations and content categories, and a library of controlled temporal stressors for applying temporal transformations.
    (b) \textit{Evaluation Protocols}. \VideoSTF assesses output repetition through three tests. \textit{Pervasive Testing} reveals widespread repetition across \VideoLLMs under different frame sampling settings across all three metrics. \textit{Temporal Stress Testing} shows that temporal transformations amplify repetition. \textit{Adversarial Exploitation} demonstrates that, through temporal transformations, videos with normal outputs can be efficiently induced to become repetitive with high success rates and few queries.
}
\label{fig:pipeline}
\vspace{-2mm}
\end{figure*}

\section{\VideoSTF}
\label{sec:preliminary}

\subsection{Output Repetition Formulation}
\label{subsec:repetition_formulation}

A \VideoLLM takes a video $\mathcal{V}$ as input, consisting of $T$ sampled frames: $\mathcal{V} = \{ I_1, \ldots, I_T \}$. 
Each frame $I_t$ is encoded by a visual encoder $\phi_v$ to produce frame-level features $\mathbf{Z}_t = \phi_v(I_t)$, which are then aggregated into video-level features $\mathbf{Z} = \{ \mathbf{Z}_1, \ldots, \mathbf{Z}_T \}$.
Then, a projector $\phi_p$ maps $\mathbf{Z}$ into the language model embedding space to produce projected visual tokens $\mathbf{H} = \phi_p(\mathbf{Z})$. 
The projected visual tokens $\mathbf{H}$ are subsequently
concatenated with prompt embeddings $\mathbf{Q} = \{ \mathbf{q}_1, \ldots, \mathbf{q}_L \}$, where $L$ denotes the token length of the prompt, to form a multimodal input sequence. 
Finally, the underlying LLM $\mathcal{F}$ generates the textual output conditioned on the fused sequence: $\mathbf{Y} = \mathcal{F} (\mathbf{H};\mathbf{Q})$.
Given a video input, the generated text output $\mathbf{Y}$ can be decomposed into $M$ token-level units, denoted as $\mathbf{Y} = \{ y_1, \ldots, y_M \}$, where each $y_i$ denotes a single token.
To formalize repetition, we employ three metrics based on $n$-gram~\cite{cavnar1994ngram}, capturing repetition occurrence, intensity, and severity, respectively.

\textbf{Repetition Rate (RR).}
We first quantify whether an output exhibits repetition using an $n$-gram-based \cite{cavnar1994ngram, yao2025understanding, fu2021theoretical} repetition rate. 
Specifically, given an output $\mathbf{Y}$,
an $n$-gram is defined as any contiguous subsequence $(y_i, y_{i+1}, \ldots, y_{i+n-1})$ of length $n$ in $\mathbf{Y}$.
Let $\mathcal{S}(\mathbf{Y})$ denote the multiset of all contiguous $n$-grams in $\mathbf{Y}$, and $\mathrm{count}(s)$ denote the number of occurrences of an $n$-gram $s \in \mathcal{S}(\mathbf{Y})$, we can define 
the maximum number of times any $n$-gram is repeated in the output $\mathbf{Y}$ as:
\(
\mathcal{R}(\mathbf{Y}) = \max_{s \in \mathcal{S}(\mathbf{Y})} \mathrm{count}(s),
\)
Given a set of $K$ generated outputs $\{\mathbf{Y}_k\}_{k=1}^{K}$, the dataset-level RR is defined as: 
\begin{equation}
\label{eq:rr}
\mathrm{RR} = \frac{1}{K} \sum_{k=1}^{K} \mathbb{I} \big( \mathcal{R}(\mathbf{Y}_k) > \tau \big),
\end{equation}
where $\mathbb{I}(\cdot)$ is the indicator function. We set $\tau = 1$ in our main experiments to capture the most direct repetition, and provide further analysis on the larger $\tau$ (more severe repetition) in Section~\ref{sec:analysis}. Thus, $\mathrm{RR}$ measures the fraction of outputs that contain at least one repeated $n$-gram.

\textbf{Repetition Intensity (RI).}
While $\mathrm{RR}$ measures whether repetition occurs, it does not capture the extent of duplication within an output. We measure repetition intensity based on $\mathrm{Rep}\text{-}n$ metric~\cite{fu2021theoretical,welleck2020neural}, which quantifies the proportion of duplicated $n$-grams: 
\begin{equation}
\mathrm{Rep}\text{-}n(\mathbf{Y})
= 1 - \frac{\lvert \mathrm{unique} \ n \text{-}\mathrm{grams}(\mathbf{Y},n) \rvert}{M-n+1}.
\end{equation}
where $M-n+1$ denotes the total number of contiguous $n$-grams in the sequence. We assess RI at the dataset-level by averaging $\mathrm{Rep}\text{-}n$ over all $K$ outputs:
\begin{equation}
\label{eq:ri}
\mathrm{RI}
= \frac{1}{K} \sum_{k=1}^{K} \mathrm{Rep}\text{-}n(\mathbf{Y}_k).
\end{equation}

\textbf{Information Entropy (IE).}
Finally, we adopt an entropy-based metric~\cite{tsai2008information} that measures the diversity of token-level $n$-grams.
Given $p_i$, which denotes the empirical probability of the $i$-th $n$-gram in $\mathbf{Y}$, we define the normalized $n$-gram entropy as:
\begin{equation}
H_\mathrm{norm}(\mathbf{Y}) = - \frac{1}{\log_2 N} \sum_{i=1}^N p_i \log_2 p_i,
\end{equation}
where $N = |\mathcal{S}(\mathbf{Y})|$. Lower entropy indicates stronger repetition and less lexical diversity.
The dataset-level IE is then computed as:
\begin{equation}
\label{eq:ie}
\mathrm{IE} = \frac{1}{K} \sum_{k=1}^{K} H_\mathrm{norm}(\mathbf{Y}_k).
\end{equation}

Each metric captures a distinct aspect of repetition. RR reflects the presence of degenerate looping behavior, RI quantifies the extent of duplication once repetition emerges, and IE captures the loss of lexical diversity in repetitive generation. Together, these metrics allow us to distinguish isolated repetition artifacts from sustained degeneration, which no single metric can capture alone.

\subsection{Standardized Video Testbed}
\label{subsec:svt}

We construct a standardized video testbed by randomly sampling 10,000 videos from several widely used public video datasets, including LLaVA-Video-178K~\cite{lin2024video-llava}, NeXT-QA~\cite{xiao2021nextqa}, ActivityNetQA~\cite{yu2019activitynetqa}, and LLaVA-Hound~\cite{zhang2024llava-hound}. As illustrated in Figure~\ref{fig:pipeline}(a), the resulting collection exhibits substantial diversity in both temporal length and semantic content. Video durations range from short clips to moderately long videos of up to 180 seconds, enabling evaluation across different temporal scales, while the content spans a wide variety of everyday scenarios, with categories such as comedy, lifestyle, and sports appearing frequently. This diversity makes the testbed representative of common \VideoLLM usage scenarios and well suited for comprehensively evaluating output repetition behavior.

\subsection{Temporal Stressor Library}
\label{subsec:tsl}

The temporal stressor library is created with a set of temporal transformations that alter the temporal structure of a video while largely preserving its semantic content.
Formally, given a video sequence $\mathcal{V} = \{ I_1, \ldots, I_T \}$, a temporal transformation $\mathcal{T}$ maps $\mathcal{V}$ to a transformed video sequence:
\begin{equation}
    \mathcal{T}: \mathcal{V} \mapsto \mathcal{V}', \quad \mathcal{V}' = \{ I'_1, \ldots, I'_{T'} \},
\end{equation}
where $T'$ may differ from $T$ depending on the transformation. 
In \VideoSTF, we employ five common temporal transformations: Add, Delete, Replace, Reverse, and Shuffle that capture different ways of modifying temporal structure.
The five transformations are designed to span a controlled spectrum from \emph{local} perturbations that preserve global temporal coherence to \emph{global} perturbations that destroy it, allowing us to isolate which aspect of temporal structure interacts with output repetition. \textbf{Add} randomly selects $k$ frames $\{ I_{i_1}, \dots, I_{i_k} \}$ from $\mathcal{V}$ and inserts them at $k$ random positions $\{ q_1, \dots, q_k \}$, producing $\mathcal{V}' = \{ I'_1, \dots, I'_{T+k} \}$ in which each inserted frame $I_{i_j}$ occupies position $q_j$ and the original frames retain their relative order; this introduces local redundancy. \textbf{Delete} removes $k$ randomly chosen frames, yielding $\mathcal{V}' = \{ I'_1, \dots, I'_{T-k} \}$ consisting of the remaining frames in their original relative order, which introduces local omissions. \textbf{Replace} randomly selects $k$ frames to be replaced, and for each selected $I_{i_m}$ draws a replacement $I_{j_m}$ uniformly from the remaining frames $\mathcal{V}\setminus\{I_{i_1},\dots,I_{i_k}\}$, producing $\mathcal{V}' = \{ I'_1, \dots, I'_T \}$ that preserves the original length but introduces local inconsistencies. Together, Add, Delete, and Replace act locally and leave the global temporal trajectory largely intact. In contrast, \textbf{Reverse} inverts the order of all frames to produce $\mathcal{V}' = \{ I_T, I_{T-1}, \dots, I_1 \}$, leaving each frame's content unchanged but globally inverting the temporal progression, and \textbf{Shuffle} applies a random permutation $\pi$ so that $I'_{t} = I_{\pi(t)}$ for $t = 1, \dots, T$, preserving the set of frames while destroying any coherent ordering. Together, the five transformations cover redundancy, omission, local inconsistency, global inversion, and global disorder, providing a principled basis for stress-testing how each aspect of temporal structure affects output repetition.

\section{Experiments}
\label{sec:experiments}

\subsection{Experiment Setup}
\label{subsec:experiment_setup}

\VideoSTF tests 10 \VideoLLMs, 
including
LLaVA-Video-7B-Qwen2~\cite{zhang2024LLaVA-Video}, 
LLaVA-Video-7B-Qwen2-Video-Only~\cite{zhang2024LLaVA-Video}, 
LLaVA-NeXT-Video-7B~\cite{zhang2024llavanextvideo}, 
LLaVA-NeXT-Video-7B-DPO~\cite{zhang2024llavanextvideo}, 
LLaVA-NeXT-Video-32B-Qwen~\cite{zhang2024llavanextvideo}, 
VideoLLaMA2~\cite{cheng2024videollama2}, 
ShareGPT4Video~\cite{chen2024sharegpt4video}, 
InternVL3.5-8B~\cite{wang2025internvl3}, 
Qwen3-VL-8B-Instruct~\cite{bai2025qwen3vl}, and
Molmo2-8B~\cite{clark2026molmo2}.
All models are tested under deterministic settings, with \texttt{do\_sample} set to \texttt{False} and temperature fixed at 0.0, to avoid decoding randomness.

We adopt the three metrics: Repetition Rate (RR), Repetition Intensity (RI), and Information Entropy (IE) (\Cref{subsec:repetition_formulation}). 
For RR, we set $n=5$, which strikes a balance between sensitivity and robustness: this choice avoids overestimating repetition due to frequent function-word patterns while reliably capturing non-trivial repeated phrases indicative of pathological repetition in \VideoLLM outputs.
For RI and IE, we set $n=1$ following \cite{yao2025understanding}, as unigram-based statistics are most sensitive to the emergence of repetition, whereas larger $n$ values 
become increasingly insensitive and less effective in distinguishing repetitive from normal outputs. Nevertheless, \VideoSTF supports all metrics with arbitrary $n$, and we report results with other $n$ values in \Cref{app:different_n}. 

\subsection{Pervasive Testing}
\label{sec:study1}
Pervasive testing evaluates whether output repetition consistently arises in  \VideoLLMs with different base LLMs and under varying numbers of sampled frames. We assess 10 \VideoLLMs with various base LLMs. For each model, we consider four different numbers of sampled frames: 8, 16, 24, and 32. By default, LLaVA-NeXT-Video-7B and LLaVA-NeXT-Video-7B-DPO do not enforce a fixed number of sampled frames, leading to fewer frames for short videos. To ensure fair comparison, we enable forced frame sampling for both models in the testing.

\begin{table}[!t]  
\centering
\caption{Repetition rates (\%), repetition intensity and information entropy of mainstream \VideoLLMs.
}
\label{tab:main_results}
\scalebox{0.82}{
\def\arraystretch{.9}
\addtolength{\tabcolsep}{-1ex}
\begin{tabular}{ccccc@{\hspace{1.5em}}ccccc}
\toprule
\textbf{Model} \textbf{(LLM)} & \textbf{Frames} & \textbf{RR} & \textbf{RI} & \textbf{IE}
&
\textbf{Model} \textbf{(LLM)} & \textbf{Frames} & \textbf{RR} & \textbf{RI} & \textbf{IE} \\
\toprule

\multirow{4}{*}{\makecell[c]{LLaVA-Video-7B-Qwen2 \\ (Qwen2)}} 
& 8 & 3 & 0.32 & 0.87
&
\multirow{4}{*}{\makecell[c]{VideoLLaMA2 \\ (Mistral-7B-Instruct-v0.2)}}
& 8 & 13 & 0.37 & 0.84 \\
& 16 & 5 & 0.33 & 0.86
&
& 16 & 15 & 0.38 & 0.84 \\
& 24 & 10 & 0.34 & 0.86
&
& 24 & 16 & 0.39 & 0.83 \\
& 32 & 7 & 0.32 & 0.86
&
& 32 & 15 & 0.38 & 0.83 \\
\midrule

\multirow{4}{*}{\makecell[c]{LLaVA-Video-7B-Qwen2-Video-Only \\ (Qwen2)}} 
& 8 & 63 & 0.47 & 0.80
&
\multirow{4}{*}{\makecell[c]{ShareGPT4Video \\ (Meta-Llama-3-8B-Instruct)}}
& 8 & 91 & 0.58 & 0.75 \\
& 16 & 67 & 0.48 & 0.80
&
& 16 & 85 & 0.57 & 0.76 \\
& 24 & 61 & 0.46 & 0.80
&
& 24 & 79 & 0.55 & 0.76 \\
& 32 & 58 & 0.46 & 0.81
&
& 32 & 82 & 0.55 & 0.77 \\
\midrule

\multirow{4}{*}{\makecell[c]{LLaVA-NeXT-Video-7B \\ (Vicuna-7b-v1.5)}}
& 8 & 15 & 0.42 & 0.82
&
\multirow{4}{*}{\makecell[c]{InternVL3.5-8B \\ (Qwen3-8B)}}
& 8 & 9 & 0.36 & 0.84 \\
& 16 & 17 & 0.41 & 0.82
&
& 16 & 7 & 0.32 & 0.86 \\
& 24 & 18 & 0.42 & 0.83
&
& 24 & 11 & 0.37 & 0.84 \\
& 32 & - & - & -
&
& 32 & 7 & 0.35 & 0.85 \\
\midrule

\multirow{4}{*}{\makecell[c]{LLaVA-NeXT-Video-7B-DPO \\ (Vicuna-7b-v1.5)}}
& 8 & 11 & 0.39 & 0.83
&
\multirow{4}{*}{\makecell[c]{Qwen3-VL-8B-Instruct \\ (Qwen3-8B)}}
& 8 & 8 & 0.36 & 0.85 \\
& 16 & 8 & 0.37 & 0.85
&
& 16 & 13 & 0.36 & 0.85 \\
& 24 & 9 & 0.38 & 0.84
&
& 24 & 11 & 0.35 & 0.85 \\
& 32 & - & - & -
&
& 32 & 11 & 0.35 & 0.84 \\
\midrule

\multirow{4}{*}{\makecell[c]{LLaVA-NeXT-Video-32B-Qwen \\ (Qwen1.5-32B)}}
& 8 & 18 & 0.35 & 0.85
&
\multirow{4}{*}{\makecell[c]{Molmo2-8B \\ (Qwen3-8B)}}
& 8 & 65 & 0.49 & 0.80 \\
& 16 & 25 & 0.36 & 0.85
&
& 16 & 69 & 0.50 & 0.79 \\
& 24 & 22 & 0.36 & 0.85
&
& 24 & 66 & 0.49 & 0.79 \\
& 32 & 29 & 0.37 & 0.84
&
& 32 & 72 & 0.51 & 0.78 \\
\bottomrule
\end{tabular}
}
\vspace{-5mm}
\end{table}

Table~\ref{tab:main_results} shows that all the 10 \VideoLLMs, despite being built on different base LLMs, exhibit output repetition (typical output repetition examples are provided in \Cref{app:pervasive_repetition}). Among them, 
ShareGPT4Video and Molmo2-8B display the most severe repetition, with repetition rates exceeding 79\% and 65\%, respectively. 
Moreover, across all models, output repetition remains stable across 8, 16, 24, and 32 sampled frames, indicating that this failure mode is insensitive to the commonly used input temporal length.
For LLaVA-NeXT-Video-7B and LLaVA-NeXT-Video-7B-DPO, we observe empty outputs when the number of sampled frames reaches 32 because the resulting visual embeddings exceed the maximum input capacity of the underlying LLMs, breaking the generation and preventing repetition from being measured.
In summary, these results indicate that output repetition is pervasive in modern \VideoLLMs on unperturbed videos, neither tied to a particular underlying LLM nor sensitive to the number of sampled frames within this range.

\begin{figure*}[t]
    \centering
    \includegraphics[width=0.98\linewidth]{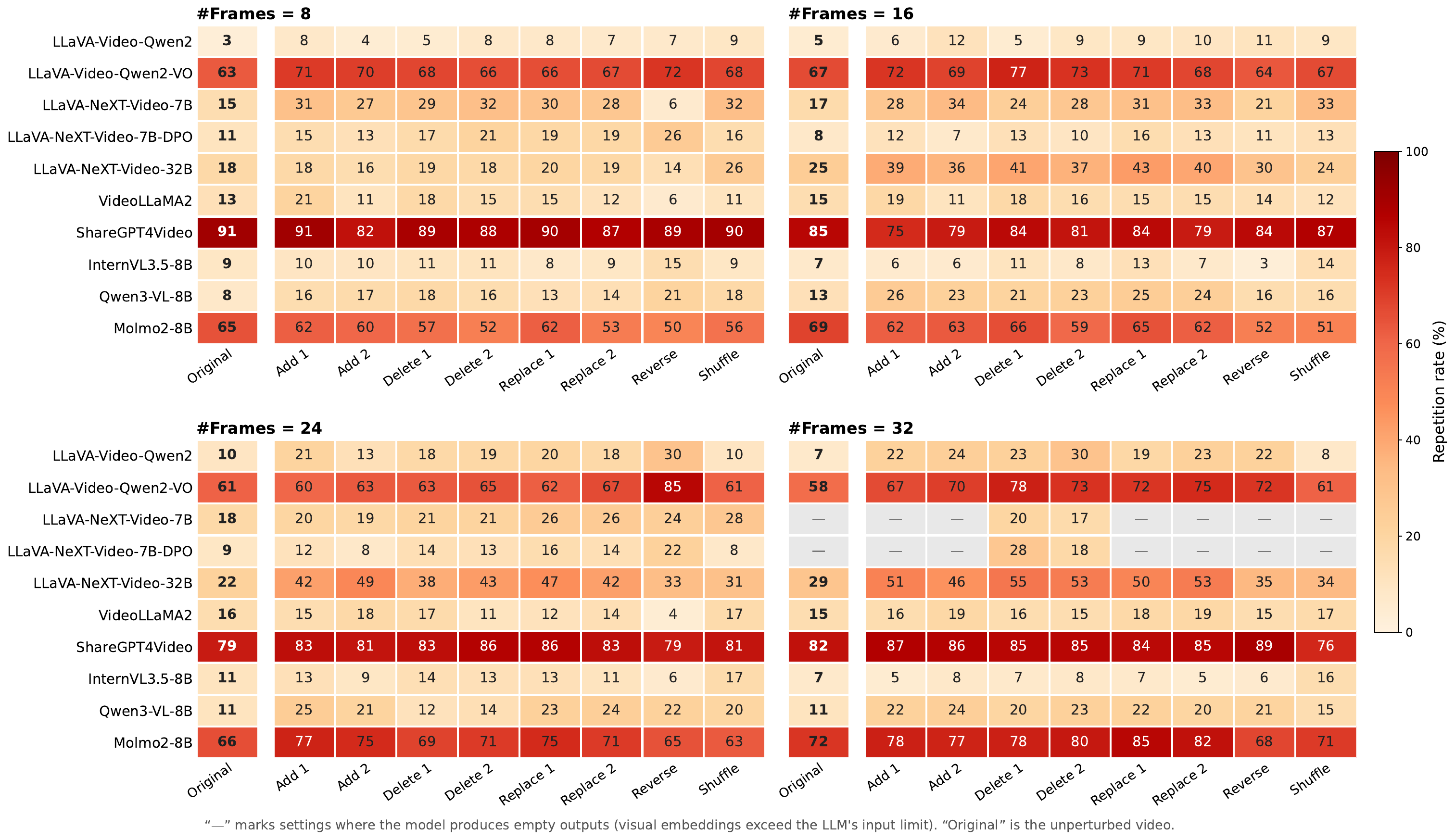}
    \vspace{-2mm}
    \caption{Repetition rates (\%) on original videos and under temporal transformations.}
    \label{fig:results_temporal_transformations}
    \vspace{-4mm}
\end{figure*}

\subsection{Temporal Stress Testing}
\label{sec:study2}

In this testing, we consider five temporal transformations, Add Frames, Delete Frames, Replace Frames, Reverse, and Shuffle (\Cref{subsec:tsl}).
For the former three ones, we evaluate two variants with $k=1$ and $k=2$, resulting in six variants: Add 1, Add 2, Delete 1, Delete 2, Replace 1, and Replace 2.
When deleting one frame, we exhaustively traverse all sampled frames and remove each frame once. When deleting two frames, we run 30 trials and randomly select two distinct frames to remove in each trial (to balance coverage across different sampled-frame numbers). 
For Add Frames and Replace Frames, where insertion or replacement positions are stochastic, we perform 30 trials. Finally, for Shuffle, we generate 30 random permutations per video.

\begin{wrapfigure}[17]{r}{7.2cm}
    \centering
    \vspace{-4mm}
    \includegraphics[width=0.99\linewidth]{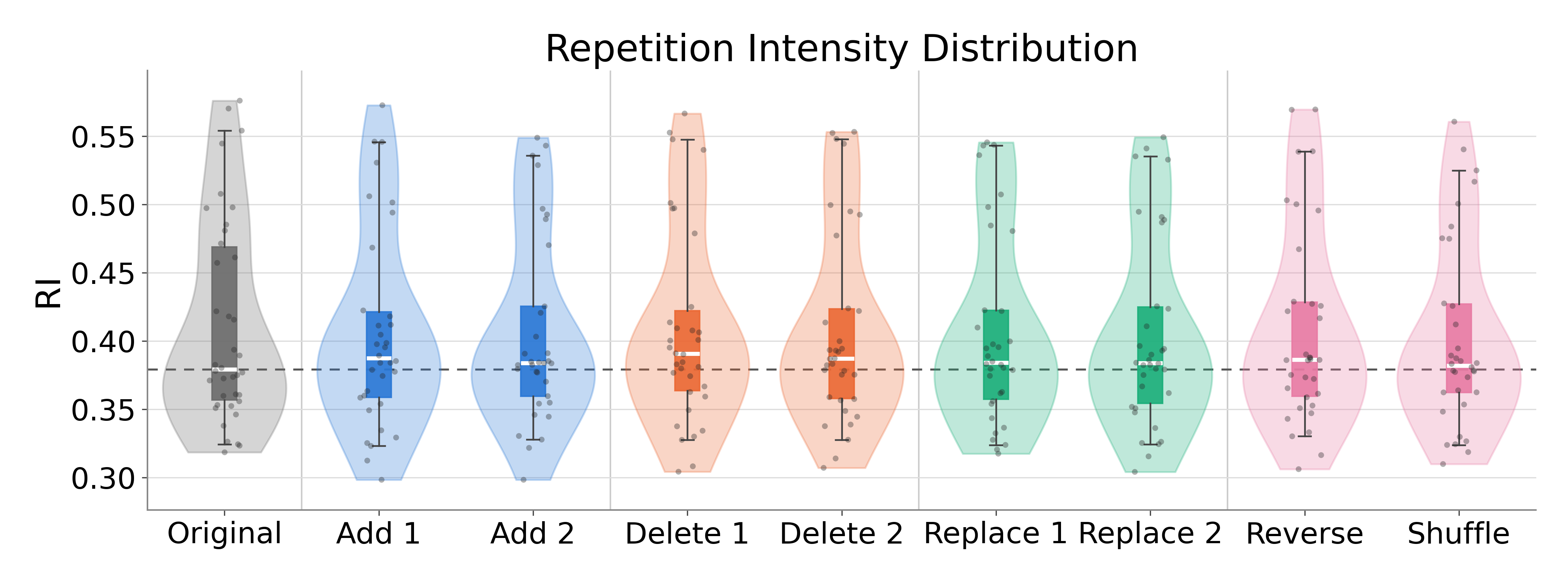}
    \includegraphics[width=0.99\linewidth]{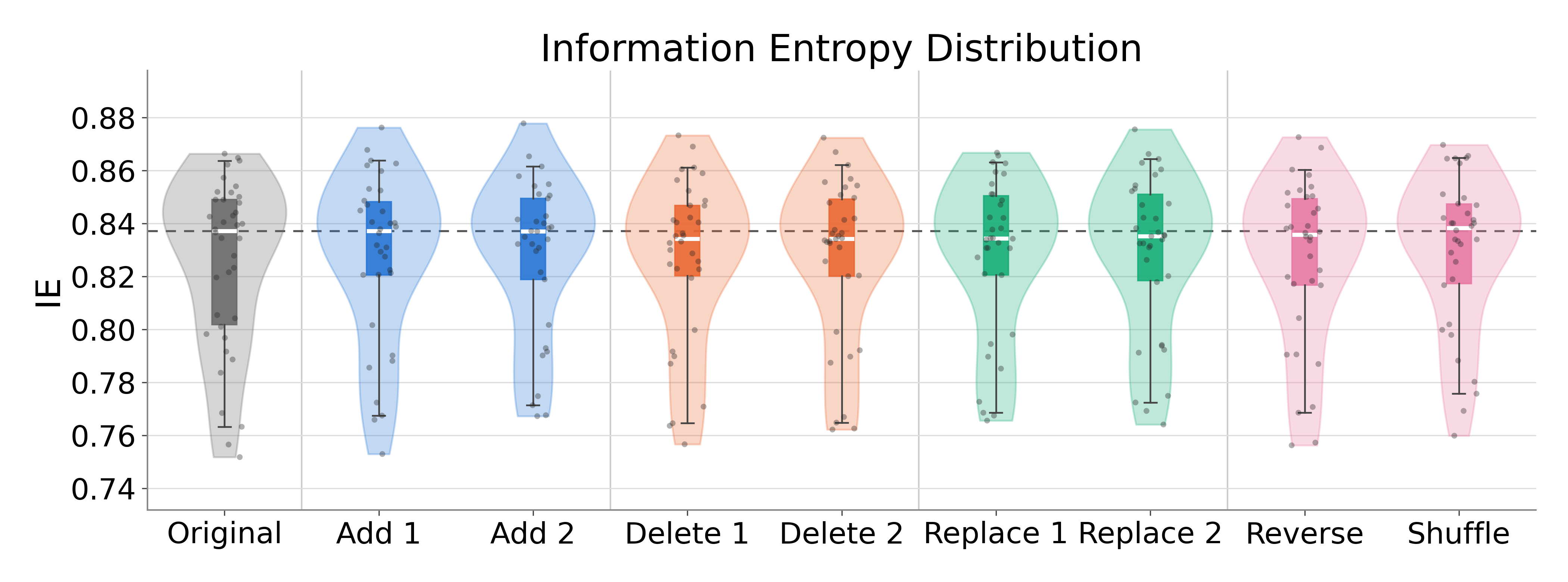}
    \caption{RI and IE distributions under original and temporally transformed inputs across \VideoLLMs.}
    \label{fig:other_metrics}
\end{wrapfigure}

Figure~\ref{fig:results_temporal_transformations} shows that temporal transformations increase repetition rates in almost all cases. We summarize this effect by computing, for each sampled frame count, the gap between the mean repetition rate over the eight stressor variants and that of the original videos, normalized by the latter. The relative increase is 133\%, 78\%, 86\%, and 205\% at 8, 16, 24, and 32 frames for LLaVA-Video-7B-Qwen2, and 108\%, 67\%, 83\%, and 90\% for Qwen3-VL-8B-Instruct, with marginal decreases in only a few cases. As shown in \Cref{fig:other_metrics}, RI and IE exhibit similar trends. Representative examples are provided in \Cref{app:transformation_repetition}. 
These results show that temporal modifications can substantially amplify repetitive generation. Since prevalence is established on unperturbed videos in \Cref{sec:study1}, this test instead measures sensitivity to temporal change. As in pervasive testing, the number of sampled frames has limited overall impact. A small subset of models (\eg LLaVA-Video-7B-Qwen2) show increasing repetition as more frames are sampled, suggesting greater susceptibility under denser sequences of visually similar frames.

\begin{figure*}[t]
    \centering
    \includegraphics[width=0.99\linewidth]{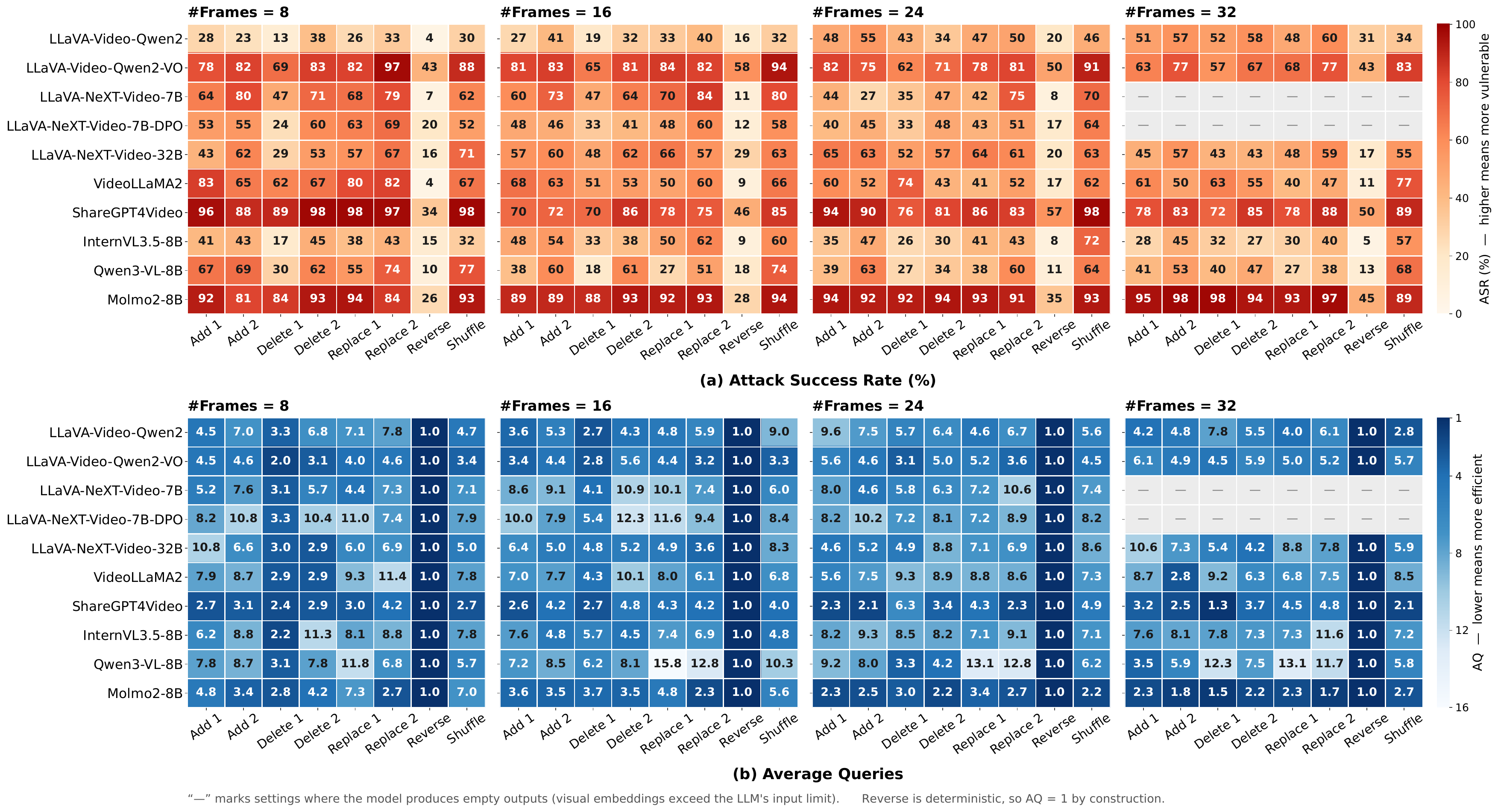}
    \vspace{-2mm}
    \caption{Attack performance of stressors on videos that originally produce non-repetitive outputs.}
    \label{fig:results_attack}
    \vspace{-4mm}
\end{figure*}

When comparing different temporal transformations, we observe a clear pattern. Add, Delete, and Replace preserve partial temporal coherence while introducing redundancy or local inconsistencies, which can reinforce similar predictions and promote repetition.  Reverse instead disrupts temporal order globally, weakening learned temporal priors and reducing repetitive descriptions. Shuffle lies between these extremes: although it destroys global order, its local randomness rarely produces fully inverted sequences, leading to fragmented but still repetition-prone cues. Because Add and Delete also alter frame count, we compare them with count-matched uniform re-sampling. Add remains stronger than its control, whereas Delete does not, indicating that part of the Delete effect is attributable to frame count rather than temporal disruption. \Cref{app:corruption} provides more details.

\subsection{Adversarial Exploitation}
\label{sec:study3}

Our temporal stress testing shows that temporal transformations can amplify output repetition in \VideoLLMs. We therefore examine whether input-level temporal manipulations can flip videos that originally produce non-repetitive outputs into ones that trigger repetition.
We consider a black-box setting, as in prior attacks on video models~\cite{jiang2019black,xie2022universal,cao2023stylefool,tang2026query} and \VideoLLMs~\cite{cao2025poisoning, liu2025protecting}, in which an attacker can only query the target \VideoLLM and apply temporal transformations to the input videos to induce repetition, 
without access to the model's parameters or internal architecture.
For each \VideoLLM, we select the video examples from pervasive testing (\Cref{sec:study1}) that originally produce normal outputs as the attack set. We leverage the temporal transformations defined in \Cref{subsec:tsl} and treat each of them as a distinct attack type. 
Given a video, we iteratively apply temporal transformations and query the target \VideoLLM with each transformed video until output repetition is observed or a preset maximum of 30 query attempts is reached.
We evaluate attack effectiveness using two metrics. 
Attack Success Rate (ASR) measures the proportion of videos whose outputs become repetitive (measured by RR), and Average Queries (AQ) reports the average number of queries required to induce repetition among successfully attacked samples.

Figure~\ref{fig:results_attack} shows that for almost all transformations, the attacks succeed with very few queries. 
The highest ASR reaches 98\%, and the maximum AQ remains as low as 15.8.
Specifically, Reverse yields the weakest attack performance due to its single and deterministic temporal change. 
In contrast, the remaining attack types achieve consistently high ASRs.
Even on models that exhibit relatively low repetition on original videos, attacks remain highly effective, with VideoLLaMA2 achieving 40\% to 83\% ASR and InternVL3.5-8B achieving 28\% to 72\% ASR.
These results indicate that output repetition is not a corner case confined to a small subset of videos. 
Overall, videos that initially produce normal outputs often fail to preserve this behavior under temporal manipulations, revealing a practical vulnerability: with only black-box query access, repetitive degeneration can be induced and, in extreme cases, driven until the output token limit is reached, leading to denial-of-service (DoS).

\section{Analysis}
\label{sec:analysis}

\Cref{sec:study1,sec:study2,sec:study3} establish that repetition is pervasive, temporally amplified, and exploitable. We now use \VideoSTF to examine \emph{why} it emerges and what it costs. 
Analyses that reuse the settings of the main experiments run on the full testbed, while the additional controls run on an 800-video subset of the testbed, so their absolute rates differ slightly from those of the main experiments.
Please refer to \Cref{app:analysis_details} for more results and detailed analysis, where \Cref{app:video_types} further shows that repetition persists on other video distributions.

\noindent\textbf{Severity Spectrum.}
Severe repetition (token cap reached) is concentrated in the most affected models (\eg ShareGPT4Video, Molmo2-8B), while ordinary repetition dominates and already degrades quality, and \Cref{fig:severity} shows that the maximum $n$-gram count thins rapidly beyond $\tau\!>\!1$. Most repetitions are thus moderate rather than DoS-level, and \VideoSTF captures the \emph{onset} of repetition.

\noindent\textbf{Repetition and Output Quality.}
On unperturbed videos in the subset, repetition tracks the information density of a description (distinct content words per 100 words) almost monotonically, with Spearman correlations of -0.95 for RI and 0.92 for IE. Severe outputs are 2.6$\times$ longer than benign ones yet contain no more distinct content words, while CLIPScore~\cite{hessel2021clipscore} stays flat when averaged over the sentences of a description. Multiple-choice accuracy on the same NExT-QA~\cite{xiao2021nextqa} clips barely depends on whether the model's description repeats (86.6\% \vs 83.6\% for LLaVA-Video-7B-Qwen2 and 84.7\% \vs 84.9\% for its Video-Only variant). This suggests that repetition is a failure of generation stability rather than of video understanding, which accuracy-based benchmarks~\cite{ning2025video, li2024mvbench, li2025benchmark} do not capture, and \Cref{app:quality} reports the full results.

\begin{table}[t]
\centering
\caption{Repetition rates (\%) under visual-input controls with fixed weights, prompt, and decoding.}
\label{tab:controls_main}
\small
\scalebox{0.92}{
\def\arraystretch{.9}
\addtolength{\tabcolsep}{-1ex}
\begin{tabular}{lccccccccccc}
\toprule
\multirow{2}{*}{Model} & \multirow{2}{*}{Text only} & \multirow{2}{*}{Black} & \multirow{2}{*}{Noise} & \multirow{2}{*}{Static} & \multicolumn{7}{c}{Genuine frames} \\
\cmidrule(lr){6-12}
 & & & & & 1 & 2 & 4 & 8 & 16 & 24 & 32 \\
\midrule
LLaVA-Video-7B-Qwen2 & 56 & 2 & 0 & 0 & 9 & 2 & 10 & 3 & 7 & 11 & 8 \\
LLaVA-Video-7B-Qwen2-Video-Only & 85 & 11 & 0 & 30 & 4 & 23 & 50 & 64 & 68 & 64 & 61 \\
ShareGPT4Video & 0 & 0 & 0 & 12 & 26 & 57 & 76 & 87 & 85 & 80 & 84 \\
Qwen3-VL-8B-Instruct & 0 & 0 & 0 & 11 & 14 & 8 & 6 & 9 & 12 & 12 & 10 \\
Molmo2-8B & 0 & 0 & 2 & 27 & 15 & 22 & 21 & 59 & 63 & 60 & 74 \\
\bottomrule
\end{tabular}
}
\vspace{-3mm}
\end{table}

\noindent\textbf{Visual-Input Controls.}
We isolate the effect of visual input by fixing the weights, prompt, and decoding configuration while varying only the video input. As summarized in \Cref{tab:controls_main}, uninformative inputs produce little repetition (0\%--11\%), while a maximally redundant static video yields less repetition than 16 genuine frames (0\%--30\% \vs 7\%--85\%). Repetition also increases with frame count only for susceptible models and largely plateaus beyond 8 frames. Although the two LLaVA-Video variants can loop without video by fabricating frame-by-frame narrations, text-only repetition does not predict video repetition across models. Likewise, the shared Qwen3-8B~\cite{yang2025qwen3} backbone does not determine the behavior of Qwen3-VL-8B-Instruct and Molmo2-8B. Together, these controls show that repetition depends on visual input but is not explained by visual redundancy alone. Instead, it increases with distinct frames to describe, while repeated spans typically re-identify the same people, objects, or scenes. Full analyses are provided in \Cref{app:controls,app:qualitative}.

\begin{figure}[t]
    \centering
    \includegraphics[width=0.84\linewidth]{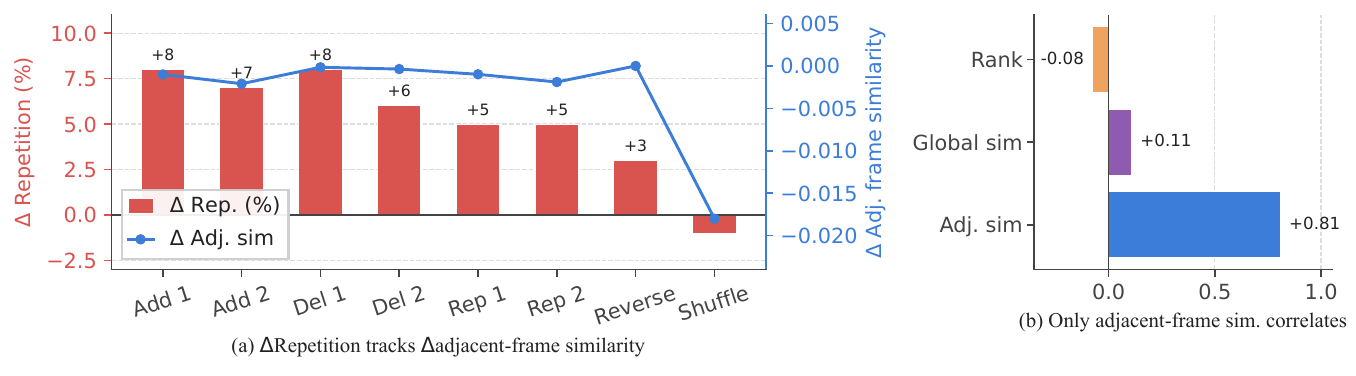}
    \vspace{-2mm}
    \caption{Feature-level changes induced by temporal transformations. 
    }
    \label{fig:feature_corr}
    \vspace{-5mm}
\end{figure}

\noindent\textbf{Local Temporal Disruption.}
Stressor-induced amplification is not explained by generic input corruption. On LLaVA-Video-7B-Qwen2 at 32 frames in the subset, replacing 2 frames with frames from a different video raises repetition from 8\% to 17\%, whereas blurring, noising, JPEG-compressing, or masking the same 2 frames changes it by at most 2\%, and a 2$\times$ speed re-sample that keeps the real order lowers it to 4\%, as detailed in \Cref{app:corruption}. Inside the encoder, \Cref{fig:feature_corr}(a) shows that $\Delta$repetition closely tracks $\Delta$adjacent-frame similarity across stressors, while global token similarity and effective rank remain essentially unchanged. \Cref{fig:feature_corr}(b) quantifies this, and only adjacent-frame similarity is strongly correlated with $\Delta$repetition (Pearson 0.81 over 8 stressors), whereas global similarity and rank are near zero. 

We therefore identify disrupted short-range temporal structure as the feature-level signal most consistently associated with repetition amplification, providing evidence of a strong relationship between local temporal disruption and repetitive behavior. 
On the decoder side, measured on the subset, the share of attention on visual tokens halves from the first to the last 50 generated tokens, and it does so equally for looping and benign outputs, so a loop does not arise because the decoder stops attending to the video, as shown in \Cref{app:attention}.

\begin{figure}[t]
    \centering
    \includegraphics[width=0.88\linewidth]{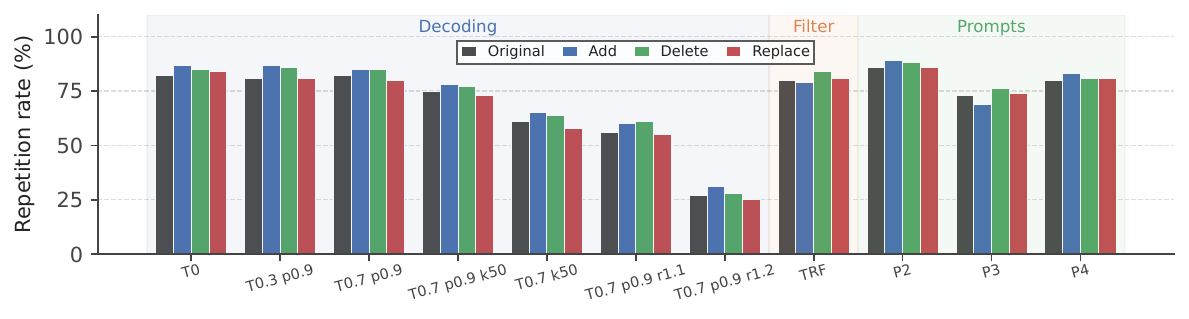}
    \vspace{-2mm}
    \caption{Repetition rates of ShareGPT4Video under decoding strategies, TRF, and prompts.}
    \label{fig:decoding_robust}
    \vspace{-5mm}
\end{figure}

\noindent\textbf{Training Data.}
An audit of 120,000 captions from six public training corpora shows that the looping vocabulary is learned, since ``continues to'' appears 41,688 times in 20,000 LLaVA-Video-178K captions, and \Cref{app:training_data} reports the full audit. Corpus repetitiveness alone does not predict model repetitiveness, as PixMo-Cap~\cite{clark2026molmo2} has the lowest caption RI yet trains the highly repetitive Molmo2-8B. The training mixture helps reconcile this, because removing non-video data from the mixture of LLaVA-Video-7B-Qwen2 moves RI from 0.33 to 0.48, toward the caption RI of 0.677. With the lower repetition of the DPO variant, this indicates that training-stage choices influence the rate, although confirming the link would require retraining with controlled caption length.

\noindent\textbf{Decoding-Time Mitigation.}
\Cref{fig:decoding_robust} reports ShareGPT4Video at 32 frames on the full testbed and shows that sampling, temporal redundancy filtering (TRF), and rephrased prompts (P2 to P4, described in \Cref{app:decoding}) have limited effect, while temporal stressors amplify repetition under \emph{every} setting. The repetition penalty is the exception. It lowers repetition from 82\% to 56\% and 27\% at penalties of 1.1 and 1.2. To quantify what this costs, we repeat the experiment on LLaVA-Video-7B-Qwen2-Video-Only in the subset and add a stronger penalty of 1.3. Repetition falls from 61\% under greedy decoding to 36\% and 10\% at penalties of 1.2 and 1.3, but CLIPScore drops by 1.7\%, 2.7\%, and 5.5\% at penalties of 1.1, 1.2, and 1.3 while output length does not fall. Multiple-choice and short-answer accuracy stay unchanged across penalty strengths because the answers are too short to repeat, as reported in \Cref{app:decoding_quality}. The penalty is thus an effective symptom-level control whose grounding cost must be tuned, and no setting we test reaches both low repetition and the original grounding. As a test-time remedy that keeps the weights, decoding, and frame budget, re-querying only the outputs flagged by RR with alternative temporal views recovers 41\% of repetitive outputs, as described in \Cref{app:requery}.

\section{Conclusion}
We present \VideoSTF, a benchmarking framework for output repetition in \VideoLLMs, comprising three $n$-gram-based metrics, a standardized testbed, and a library of controlled temporal stressors. Together with the analyses it enables, \VideoSTF identifies output repetition as a pervasive, temporally amplified, and exploitable failure mode that is not explained by visual redundancy, and whose amplification tracks local temporal disruption. Repetition penalties are the only common mitigation that lowers repetition, but the gain comes with weaker visual grounding, so decoding-time control alone is not a satisfactory remedy. Our training-data audit instead points to video-caption supervision as a promising target for mitigation. Limitations and broader impacts are discussed in \Cref{app:limitations}. We hope \VideoSTF serves as a foundation on which the community builds toward more robust, temporally stable video-language systems.

\section*{Acknowledgments}
We thank the reviewers for their constructive comments. This paper was supported in part by National University of Singapore and University of New South Wales.

{
\small

\bibliography{egbib}

@article{wang2024qwen2,
  title={Qwen2-vl: Enhancing vision-language model's perception of the world at any resolution},
  author={Wang, Peng and Bai, Shuai and Tan, Sinan and Wang, Shijie and Fan, Zhihao and Bai, Jinze and Chen, Keqin and Liu, Xuejing and Wang, Jialin and Ge, Wenbin and others},
  journal={arXiv preprint arXiv:2409.12191},
  year={2024}
}

@article{chen2024InternVL2.5,
  title={Expanding performance boundaries of open-source multimodal models with model, data, and test-time scaling},
  author={Chen, Zhe and Wang, Weiyun and Cao, Yue and Liu, Yangzhou and Gao, Zhangwei and Cui, Erfei and Zhu, Jinguo and Ye, Shenglong and Tian, Hao and Liu, Zhaoyang and others},
  journal={arXiv preprint arXiv:2412.05271},
  year={2024}
}

@article{wang2025internvl3,
  title={Internvl3. 5: Advancing open-source multimodal models in versatility, reasoning, and efficiency},
  author={Wang, Weiyun and Gao, Zhangwei and Gu, Lixin and Pu, Hengjun and Cui, Long and Wei, Xingguang and Liu, Zhaoyang and Jing, Linglin and Ye, Shenglong and Shao, Jie and others},
  journal={arXiv preprint arXiv:2508.18265},
  year={2025}
}

@inproceedings{yang2023vid2seq,
  title={Vid2seq: Large-scale pretraining of a visual language model for dense video captioning},
  author={Yang, Antoine and Nagrani, Arsha and Seo, Paul Hongsuck and Miech, Antoine and Pont-Tuset, Jordi and Laptev, Ivan and Sivic, Josef and Schmid, Cordelia},
  booktitle={Proceedings of the IEEE/CVF conference on computer vision and pattern recognition},
  pages={10714--10726},
  year={2023}
}

@article{chen2024sharegpt4video,
  title={Sharegpt4video: Improving video understanding and generation with better captions},
  author={Chen, Lin and Wei, Xilin and Li, Jinsong and Dong, Xiaoyi and Zhang, Pan and Zang, Yuhang and Chen, Zehui and Duan, Haodong and Tang, Zhenyu and Yuan, Li and others},
  journal={Advances in Neural Information Processing Systems},
  volume={37},
  pages={19472--19495},
  year={2024}
}

@article{li2023videochat,
  title={Videochat: Chat-centric video understanding},
  author={Li, KunChang and He, Yinan and Wang, Yi and Li, Yizhuo and Wang, Wenhai and Luo, Ping and Wang, Yali and Wang, Limin and Qiao, Yu},
  journal={arXiv preprint arXiv:2305.06355},
  year={2023}
}

@inproceedings{tang2025aks,
  title={Adaptive keyframe sampling for long video understanding},
  author={Tang, Xi and Qiu, Jihao and Xie, Lingxi and Tian, Yunjie and Jiao, Jianbin and Ye, Qixiang},
  booktitle={Proceedings of the Computer Vision and Pattern Recognition Conference},
  pages={29118--29128},
  year={2025}
}

@inproceedings{gao2024inducing,
  title={Inducing High Energy-Latency of Large Vision-Language Models with Verbose Images},
  author={Gao, Kuofeng and Bai, Yang and Gu, Jindong and Xia, Shu-Tao and Torr, Philip and Li, Zhifeng and Liu, Wei},
  booktitle={Proceedings of the International Conference on Learning Representations},
  year={2024}
}

@article{ning2025video,
  title={Video-bench: A comprehensive benchmark and toolkit for evaluating video-based large language models},
  author={Ning, Munan and Zhu, Bin and Xie, Yujia and Lin, Bin and Cui, Jiaxi and Yuan, Lu and Chen, Dongdong and Yuan, Li},
  journal={Computational Visual Media},
  year={2025},
  publisher={TUP}
}

@inproceedings{li2024mvbench,
  title={Mvbench: A comprehensive multi-modal video understanding benchmark},
  author={Li, Kunchang and Wang, Yali and He, Yinan and Li, Yizhuo and Wang, Yi and Liu, Yi and Wang, Zun and Xu, Jilan and Chen, Guo and Luo, Ping and others},
  booktitle={Proceedings of the IEEE/CVF Conference on Computer Vision and Pattern Recognition},
  pages={22195--22206},
  year={2024}
}

@article{li2025benchmark,
  title={Benchmark evaluations, applications, and challenges of large vision language models: A survey},
  author={Li, Zongxia and Wu, Xiyang and Du, Hongyang and Nghiem, Huy and Shi, Guangyao},
  journal={arXiv preprint arXiv:2501.02189},
  volume={1},
  year={2025}
}

@article{wu2024longvideobench,
  title={Longvideobench: A benchmark for long-context interleaved video-language understanding},
  author={Wu, Haoning and Li, Dongxu and Chen, Bei and Li, Junnan},
  journal={Advances in Neural Information Processing Systems},
  volume={37},
  pages={28828--28857},
  year={2024}
}

@article{cao2025video,
  title={Video simpleqa: Towards factuality evaluation in large video language models},
  author={Cao, Meng and Hu, Pengfei and Wang, Yingyao and Gu, Jihao and Tang, Haoran and Zhao, Haoze and Wang, Chen and Dong, Jiahua and Yu, Wangbo and Zhang, Ge and others},
  journal={arXiv preprint arXiv:2503.18923},
  year={2025}
}

@inproceedings{cao2026failures,
  title={Failures to surface harmful contents in video large language models},
  author={Cao, Yuxin and Song, Wei and Wang, Derui and Xue, Jingling and Dong, Jin Song},
  booktitle={Proceedings of the AAAI Conference on Artificial Intelligence},
  volume={40},
  number={42},
  pages={35331--35339},
  year={2026}
}

@inproceedings{li2025vidhalluc,
  title={Vidhalluc: Evaluating temporal hallucinations in multimodal large language models for video understanding},
  author={Li, Chaoyu and Im, Eun Woo and Fazli, Pooyan},
  booktitle={Proceedings of the Computer Vision and Pattern Recognition Conference},
  pages={13723--13733},
  year={2025}
}

@article{zhou2025flattery,
  title={Flattery in motion: Benchmarking and analyzing sycophancy in video-llms},
  author={Zhou, Wenrui and Hendy, Mohamed and Yang, Shu and Yang, Qingsong and Guo, Zikun and Luo, Yuyu and Hu, Lijie and Wang, Di},
  journal={arXiv preprint arXiv:2506.07180},
  year={2025}
}

@article{zhang2024LLaVA-Video,
  title={Video instruction tuning with synthetic data},
  author={Zhang, Yuanhan and Wu, Jinming and Li, Wei and Li, Bo and Ma, Zejun and Liu, Ziwei and Li, Chunyuan},
  journal={arXiv preprint arXiv:2410.02713},
  year={2024}
}

@misc{zhang2024llavanextvideo,
  title={LLaVA-NeXT: A Strong Zero-shot Video Understanding Model},
  url={https://llava-vl.github.io/blog/2024-04-30-llava-next-video/},
  author={Zhang, Yuanhan and Li, Bo and Liu, haotian and Lee, Yong jae and Gui, Liangke and Fu, Di and Feng, Jiashi and Liu, Ziwei and Li, Chunyuan},
  month={April},
  year={2024}
}

@article{cheng2024videollama2,
  title={Videollama 2: Advancing spatial-temporal modeling and audio understanding in video-llms},
  author={Cheng, Zesen and Leng, Sicong and Zhang, Hang and Xin, Yifei and Li, Xin and Chen, Guanzheng and Zhu, Yongxin and Zhang, Wenqi and Luo, Ziyang and Zhao, Deli and others},
  journal={arXiv preprint arXiv:2406.07476},
  year={2024}
}

@article{bai2025qwen3vl,
  title={Qwen3-vl technical report},
  author={Bai, Shuai and Cai, Yuxuan and Chen, Ruizhe and Chen, Keqin and Chen, Xionghui and Cheng, Zesen and Deng, Lianghao and Ding, Wei and Gao, Chang and Ge, Chunjiang and others},
  journal={arXiv preprint arXiv:2511.21631},
  year={2025}
}

@article{clark2026molmo2,
  title={Molmo2: Open Weights and Data for Vision-Language Models with Video Understanding and Grounding},
  author={Clark, Christopher and Zhang, Jieyu and Ma, Zixian and Park, Jae Sung and Salehi, Mohammadreza and Tripathi, Rohun and Lee, Sangho and Ren, Zhongzheng and Kim, Chris Dongjoo and Yang, Yinuo and others},
  journal={arXiv preprint arXiv:2601.10611},
  year={2026}
}

@article{zhang2024llava-hound,
  title={Direct preference optimization of video large multimodal models from language model reward},
  author={Zhang, Ruohong and Gui, Liangke and Sun, Zhiqing and Feng, Yihao and Xu, Keyang and Zhang, Yuanhan and Fu, Di and Li, Chunyuan and Hauptmann, Alexander and Bisk, Yonatan and others},
  journal={arXiv preprint arXiv:2404.01258},
  year={2024}
}

@inproceedings{lin2024video-llava,
  title={Video-LLaVA: Learning United Visual Representation by Alignment Before Projection},
  author={Lin, Bin and Ye, Yang and Zhu, Bin and Cui, Jiaxi and Ning, Munan and Jin, Peng and Yuan, Li},
  booktitle={Proceedings of the 2024 Conference on Empirical Methods in Natural Language Processing},
  pages={5971--5984},
  year={2024}
}

@inproceedings{li2024llama-vid,
  title={Llama-vid: An image is worth 2 tokens in large language models},
  author={Li, Yanwei and Wang, Chengyao and Jia, Jiaya},
  booktitle={European Conference on Computer Vision},
  pages={323--340},
  year={2024},
  organization={Springer}
}

@article{zhang2025videollama3,
  title={VideoLLaMA 3: Frontier Multimodal Foundation Models for Image and Video Understanding},
  author={Zhang, Boqiang and Li, Kehan and Cheng, Zesen and Hu, Zhiqiang and Yuan, Yuqian and Chen, Guanzheng and Leng, Sicong and Jiang, Yuming and Zhang, Hang and Li, Xin and others},
  journal={arXiv preprint arXiv:2501.13106},
  year={2025}
}

@inproceedings{fu2021theoretical,
  title={A theoretical analysis of the repetition problem in text generation},
  author={Fu, Zihao and Lam, Wai and So, Anthony Man-Cho and Shi, Bei},
  booktitle={Proceedings of the AAAI Conference on Artificial Intelligence},
  volume={35},
  number={14},
  pages={12848--12856},
  year={2021}
}

@article{li2023repetition,
  title={Repetition in repetition out: Towards understanding neural text degeneration from the data perspective},
  author={Li, Huayang and Lan, Tian and Fu, Zihao and Cai, Deng and Liu, Lemao and Collier, Nigel and Watanabe, Taro and Su, Yixuan},
  journal={Advances in Neural Information Processing Systems},
  volume={36},
  pages={72888--72903},
  year={2023}
}

@inproceedings{yao2025understanding,
  title={Understanding the repeat curse in large language models from a feature perspective},
  author={Yao, Junchi and Yang, Shu and Xu, Jianhua and Hu, Lijie and Li, Mengdi and Wang, Di},
  booktitle={Proceedings of the Annual Meeting of the Association for Computational Linguistics},
  year={2025}
}

@inproceedings{cavnar1994ngram,
  title={N-gram-based text categorization},
  author={Cavnar, William B and Trenkle, John M and others},
  booktitle={Proceedings of SDAIR-94, 3rd annual symposium on document analysis and information retrieval},
  volume={161175},
  pages={14},
  year={1994},
  organization={Las Vegas, NV}
}

@inproceedings{welleck2020neural,
  title={Neural Text Generation With Unlikelihood Training},
  author={Welleck, Sean and Kulikov, Ilia and Roller, Stephen and Dinan, Emily and Cho, Kyunghyun and Weston, Jason},
  booktitle={Proceedings of the International Conference on Learning Representations},
  year={2020}
}

@article{tsai2008information,
  title={Information entropy measure for evaluation of image quality},
  author={Tsai, Du-Yih and Lee, Yongbum and Matsuyama, Eri},
  journal={Journal of digital imaging},
  volume={21},
  number={3},
  pages={338--347},
  year={2008},
  publisher={Springer}
}

@inproceedings{xiao2021nextqa,
  title={Next-qa: Next phase of question-answering to explaining temporal actions},
  author={Xiao, Junbin and Shang, Xindi and Yao, Angela and Chua, Tat-Seng},
  booktitle={Proceedings of the IEEE/CVF conference on computer vision and pattern recognition},
  pages={9777--9786},
  year={2021}
}

@inproceedings{yu2019activitynetqa,
  title={Activitynet-qa: A dataset for understanding complex web videos via question answering},
  author={Yu, Zhou and Xu, Dejing and Yu, Jun and Yu, Ting and Zhao, Zhou and Zhuang, Yueting and Tao, Dacheng},
  booktitle={Proceedings of the AAAI Conference on Artificial Intelligence},
  volume={33},
  number={01},
  pages={9127--9134},
  year={2019}
}

@inproceedings{jiang2019black,
  title={Black-box adversarial attacks on video recognition models},
  author={Jiang, Linxi and Ma, Xingjun and Chen, Shaoxiang and Bailey, James and Jiang, Yu-Gang},
  booktitle={Proceedings of the 27th ACM international conference on multimedia},
  pages={864--872},
  year={2019}
}

@inproceedings{xie2022universal,
  title={Universal 3-dimensional perturbations for black-box attacks on video recognition systems},
  author={Xie, Shangyu and Wang, Han and Kong, Yu and Hong, Yuan},
  booktitle={IEEE Symposium on Security and Privacy},
  pages={1390--1407},
  year={2022},
  organization={IEEE},
}

@inproceedings{cao2023stylefool,
  title={Stylefool: Fooling video classification systems via style transfer},
  author={Cao, Yuxin and Xiao, Xi and Sun, Ruoxi and Wang, Derui and Xue, Minhui and Wen, Sheng},
  booktitle={IEEE symposium on security and privacy},
  pages={1631--1648},
  year={2023},
  organization={IEEE}
}

@article{tang2026query,
  title={Query-Efficient Video Adversarial Attack with Stylized Logo on Service Computing},
  author={Tang, Duoxun and Cao, Yuxin and Xiao, Xi and Wang, Derui and Wen, Sheng and Zhu, Tianqing},
  journal={IEEE Transactions on Dependable and Secure Computing},
  year={2026},
  publisher={IEEE}
}

@article{cao2025poisoning,
  title={Poisoning Prompt-Guided Sampling in Video Large Language Models},
  author={Cao, Yuxin and Song, Wei and Xue, Jingling and Dong, Jin Song},
  journal={arXiv preprint arXiv:2509.20851},
  year={2025}
}

@inproceedings{liu2025protecting,
  title={Protecting your video content: Disrupting automated video-based llm annotations},
  author={Liu, Haitong and Gao, Kuofeng and Bai, Yang and Li, Jinmin and Shan, Jinxiao and Dai, Tao and Xia, Shu-Tao},
  booktitle={Proceedings of the Computer Vision and Pattern Recognition Conference},
  pages={24056--24065},
  year={2025}
}

@inproceedings{hessel2021clipscore,
  title={{CLIPScore}: A Reference-free Evaluation Metric for Image Captioning},
  author={Hessel, Jack and Holtzman, Ari and Forbes, Maxwell and Le Bras, Ronan and Choi, Yejin},
  booktitle={Proceedings of the 2021 Conference on Empirical Methods in Natural Language Processing},
  pages={7514--7528},
  year={2021}
}

@article{yang2025qwen3,
  title={Qwen3 technical report},
  author={Yang, An and Li, Anfeng and Yang, Baosong and Zhang, Beichen and Hui, Binyuan and Zheng, Bo and Yu, Bowen and Gao, Chang and Huang, Chengen and Lv, Chenxu and others},
  journal={arXiv preprint arXiv:2505.09388},
  year={2025}
}

@article{mann1947test,
  title={On a test of whether one of two random variables is stochastically larger than the other},
  author={Mann, Henry B and Whitney, Donald R},
  journal={The annals of mathematical statistics},
  pages={50--60},
  year={1947},
  publisher={JSTOR}
}

@article{devries2017cutout,
  title={Improved regularization of convolutional neural networks with cutout},
  author={DeVries, Terrance and Taylor, Graham W},
  journal={arXiv preprint arXiv:1708.04552},
  year={2017}
}

@inproceedings{chen2024sharegpt4v,
  title={Sharegpt4v: Improving large multi-modal models with better captions},
  author={Chen, Lin and Li, Jinsong and Dong, Xiaoyi and Zhang, Pan and He, Conghui and Wang, Jiaqi and Zhao, Feng and Lin, Dahua},
  booktitle={European Conference on Computer Vision},
  pages={370--387},
  year={2024},
  organization={Springer}
}

@article{liu2023visual,
  title={Visual instruction tuning},
  author={Liu, Haotian and Li, Chunyuan and Wu, Qingyang and Lee, Yong Jae},
  journal={Advances in neural information processing systems},
  volume={36},
  pages={34892--34916},
  year={2023}
}

@inproceedings{holtzman2020curious,
  title={The Curious Case of Neural Text Degeneration},
  author={Holtzman, Ari and Buys, Jan and Du, Li and Forbes, Maxwell and Choi, Yejin},
  booktitle={International Conference on Learning Representations},
  year={2020}
}

@inproceedings{fan2018hierarchical,
  title={Hierarchical neural story generation},
  author={Fan, Angela and Lewis, Mike and Dauphin, Yann},
  booktitle={Proceedings of the 56th Annual Meeting of the Association for Computational Linguistics},
  pages={889--898},
  year={2018}
}

@article{keskar2019ctrl,
  title={Ctrl: A conditional transformer language model for controllable generation},
  author={Keskar, Nitish Shirish and McCann, Bryan and Varshney, Lav R and Xiong, Caiming and Socher, Richard},
  journal={arXiv preprint arXiv:1909.05858},
  year={2019}
}

@inproceedings{song2025vidtoken,
  title={ViDToken: A Video-Transformer-Based Latent Token Defense for Adversarial Video Detection},
  author={Song, Wei and Sui, Yulei and Xing, Zhenchang and Zhu, Liming and Xue, Jingling},
  booktitle={International Symposium on Research in Attacks, Intrusions and Defenses},
  pages={566--582},
  year={2025},
  organization={IEEE}
}

@article{chen2026webagentguard,
  title={WebAgentGuard: A Reasoning-Driven Guard Model for Detecting Prompt Injection Attacks in Web Agents},
  author={Chen, Yulin and Cao, Tri and Li, Haoran and Liu, Yue and Li, Yibo and He, Yufei and Khoi, Le Minh and Song, Yangqiu and Yan, Shuicheng and Hooi, Bryan},
  journal={arXiv preprint arXiv:2604.12284},
  year={2026}
}

@inproceedings{chen-etal-2026-robustness,
  title={Robustness via referencing: Defending against prompt injection attacks by referencing the executed instruction},
  author={Chen, Yulin and Li, Haoran and Sui, Yuan and Liu, Yue and He, Yufei and Bai, Xiaoling and Fei, Chi and Yabo, Li and Ma, Haozhe and Song, Yangqiu and Hooi, Bryan},
  booktitle={Findings of the Association for Computational Linguistics: ACL 2026},
  pages={1194--1215},
  year={2026}
}

@inproceedings{song2024secvid,
  title={Correction-based defense against adversarial video attacks via Discretization-Enhanced video compressive sensing},
  author={Song, Wei and Cong, Cong and Zhong, Haonan and Xue, Jingling},
  booktitle={USENIX Security Symposium},
  pages={3603--3620},
  year={2024}
}

@article{robey2023smoothllm,
  title={Smoothllm: Defending large language models against jailbreaking attacks},
  author={Robey, Alexander and Wong, Eric and Hassani, Hamed and Pappas, George J},
  journal={arXiv preprint arXiv:2310.03684},
  year={2023}
}

@inproceedings{fu2025videomme,
  title={Video-mme: The first-ever comprehensive evaluation benchmark of multi-modal llms in video analysis},
  author={Fu, Chaoyou and Dai, Yuhan and Luo, Yongdong and Li, Lei and Ren, Shuhuai and Zhang, Renrui and Wang, Zihan and Zhou, Chenyu and Shen, Yunhang and Zhang, Mengdan and others},
  booktitle={Proceedings of the IEEE/CVF conference on computer vision and pattern recognition},
  pages={24108--24118},
  year={2025}
}

@inproceedings{grauman2022ego4d,
  title={Ego4d: Around the world in 3,000 hours of egocentric video},
  author={Grauman, Kristen and Westbury, Andrew and Byrne, Eugene and Chavis, Zachary and Furnari, Antonino and Girdhar, Rohit and Hamburger, Jackson and Jiang, Hao and Liu, Miao and Liu, Xingyu and others},
  booktitle={Proceedings of the IEEE/CVF conference on computer vision and pattern recognition},
  pages={18995--19012},
  year={2022}
}

@article{yi2019clevrer,
  title={Clevrer: Collision events for video representation and reasoning},
  author={Yi, Kexin and Gan, Chuang and Li, Yunzhu and Kohli, Pushmeet and Wu, Jiajun and Torralba, Antonio and Tenenbaum, Joshua B},
  journal={arXiv preprint arXiv:1910.01442},
  year={2019}
}

@article{li2025vidsme,
  title={Vid-sme: Membership inference attacks against large video understanding models},
  author={Li, Qi and Yu, Runpeng and Wang, Xinchao},
  journal={arXiv preprint arXiv:2506.03179},
  year={2025}
}
\bibliographystyle{IEEEtran}
}

\clearpage

\appendix

\begin{center}
{\LARGE\bfseries Appendix}
\end{center}
\vspace{2mm}

\section{Computational Resource}\label{app:compute}
All experiments are conducted on two RTX 4090 GPUs, which are sufficient to support the evaluated models. As our benchmark is designed to be extensible to future and more complex VideoLLMs, reproducibility should follow the computational requirements specified by each individual model.

\section{More Results on Main Experiments}\label{app:main_results}
This section presents additional results for the main experiments, covering the sensitivity of \VideoSTF to different choices of $n$, representative repetition examples across \VideoLLMs, and qualitative results under temporal stressors. To improve the layout and overall presentation of the appendix, the figures associated with \Cref{app:pervasive_repetition} and \Cref{app:transformation_repetition} are placed together at the end of the appendix, following all other sections.

\subsection{Repetition Results with Different \texorpdfstring{$n$}{n}}\label{app:different_n}
\begin{figure}[ht]
    \centering
    \includegraphics[width=0.7\linewidth]{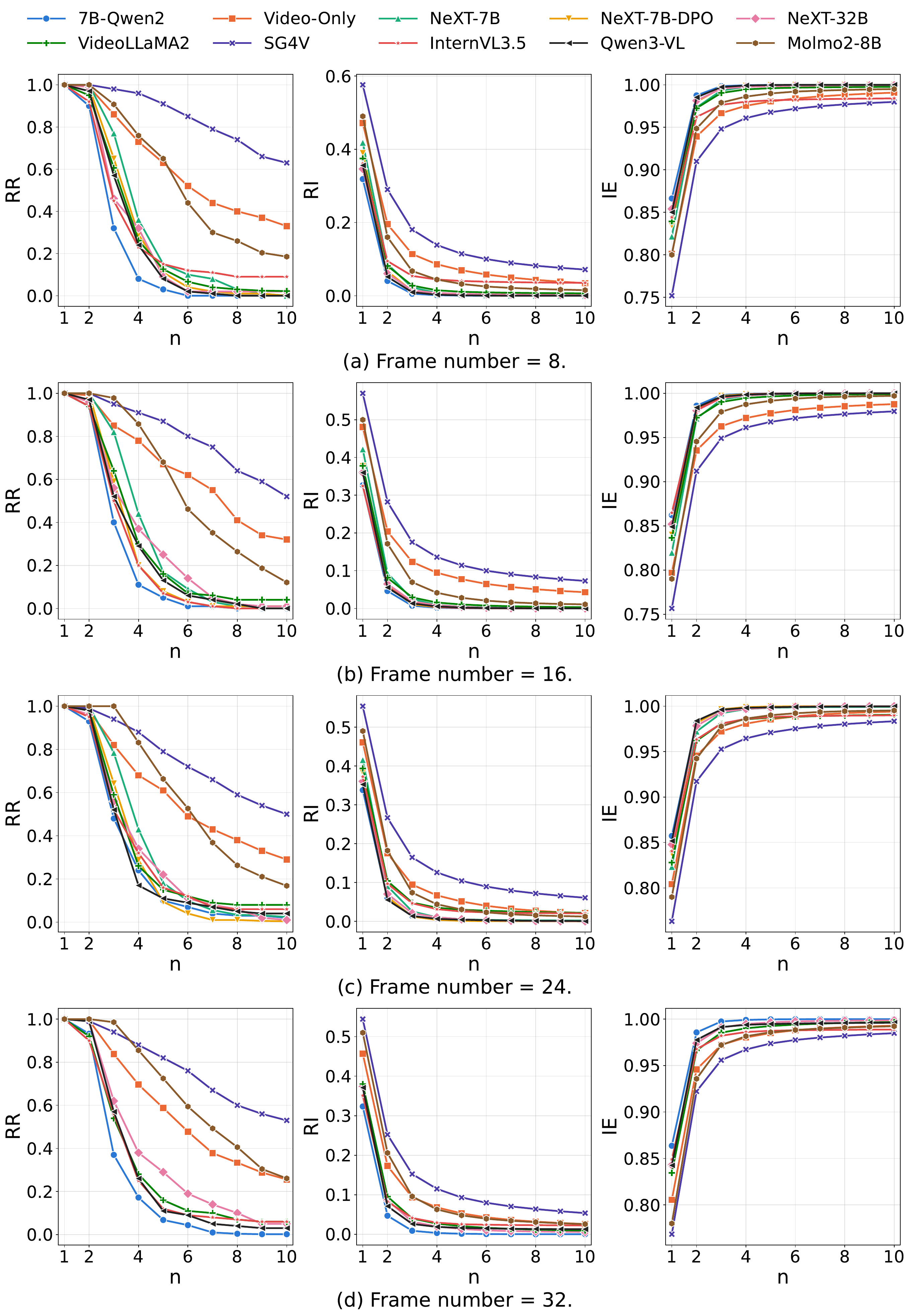}
    \caption{Repetition results of different \VideoLLMs under three metrics across varying $n$.}
    \label{fig:ablation_n}
    \vspace{-3mm}
\end{figure}

Different choices of $n$ exhibit varying sensitivity to repetition. 
We evaluate all three metrics, RR, RI, and IE (\Cref{subsec:repetition_formulation}) with $n$ ranging from 1 to 10. 
As shown in Figure~\ref{fig:ablation_n}, smaller $n$ values (\eg $n\le2$) yield higher RR but tend to overcount common short phrases, whereas larger $n$ values increasingly fail to capture repetitive patterns, especially when $n\ge7$. 
For RI and IE, repetition becomes difficult to observe when $n\ge4$.
Importantly, the relative repetition tendencies across models remain stable across different $n$ values, with ShareGPT4Video, LLaVA-Video-7B-Qwen2-Video-Only and Molmo2-8B exhibiting higher repetition. 
In general, although we use $n=5$ for RR and $n=1$ for RI and IE in the main paper, chosen to balance sensitivity and robustness for RR and to provide higher sensitivity to repetition for RI and IE, the same repetition trends and cross-model comparisons remain consistent across other choices of $n$.

\subsection{Examples of Output Repetition on Different \VideoLLMs}\label{app:pervasive_repetition}
\begin{figure}[t]
    \centering
    \includegraphics[width=0.9\linewidth]{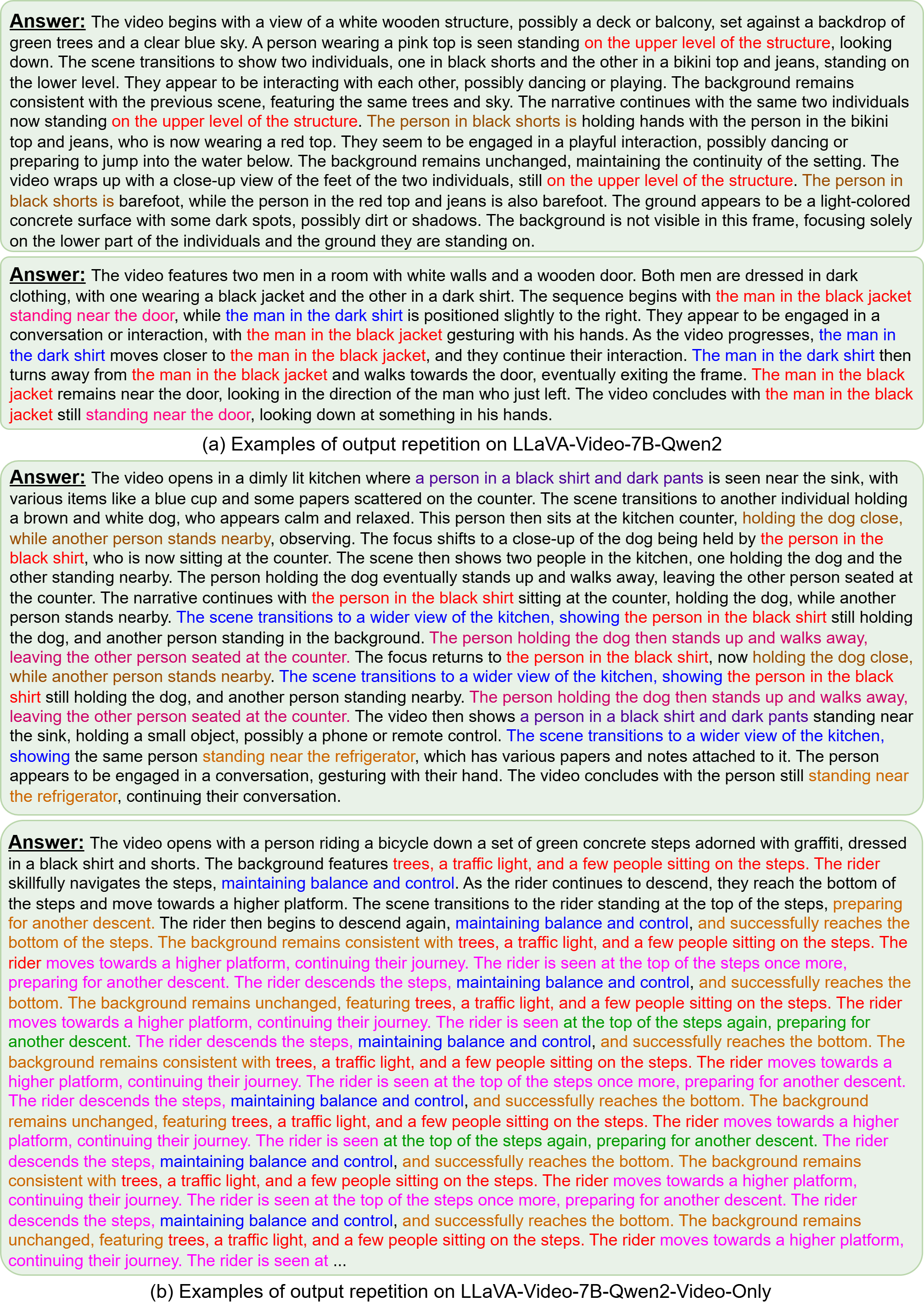}
    \caption{Examples of repetitive outputs generated by \textit{LLaVA-Video-7B-Qwen2} and \textit{LLaVA-Video-7B-Qwen2-Video-Only}, with repeated phrases or sentences highlighted in different colors.}
    \label{fig:examples12}
\end{figure}

\begin{figure}[t]
    \centering
    \includegraphics[width=0.9\linewidth]{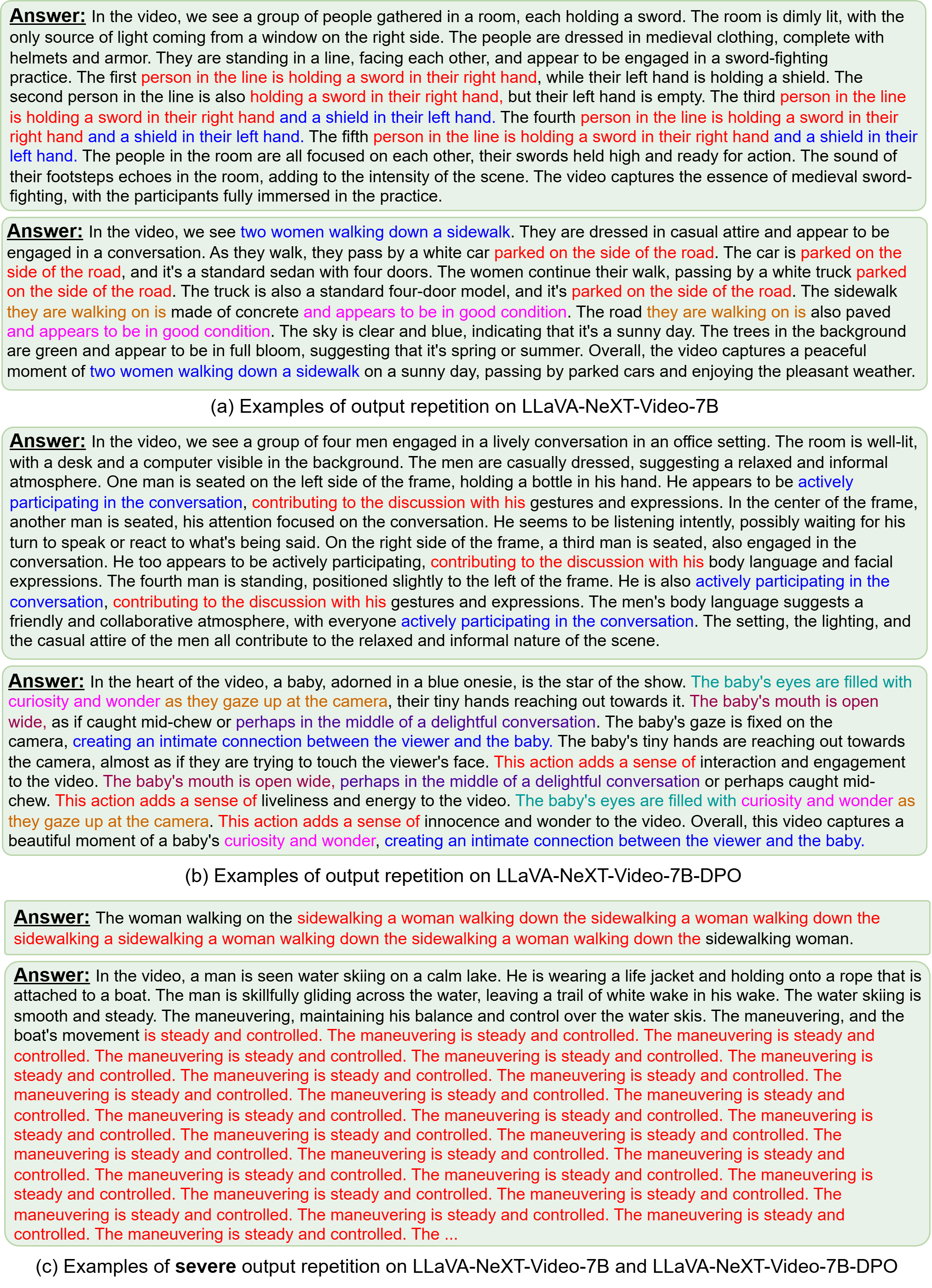}
    \caption{Examples of repetitive outputs generated by \textit{LLaVA-NeXT-Video-7B} and \textit{LLaVA-NeXT-Video-7B-DPO}, with repeated phrases or sentences highlighted in different colors. We observe highly repetitive responses when the sampled frame number falls within $[28,32)$.}
    \label{fig:examples34}
\end{figure}

\begin{figure}[t]
    \centering
    \includegraphics[width=0.9\linewidth]{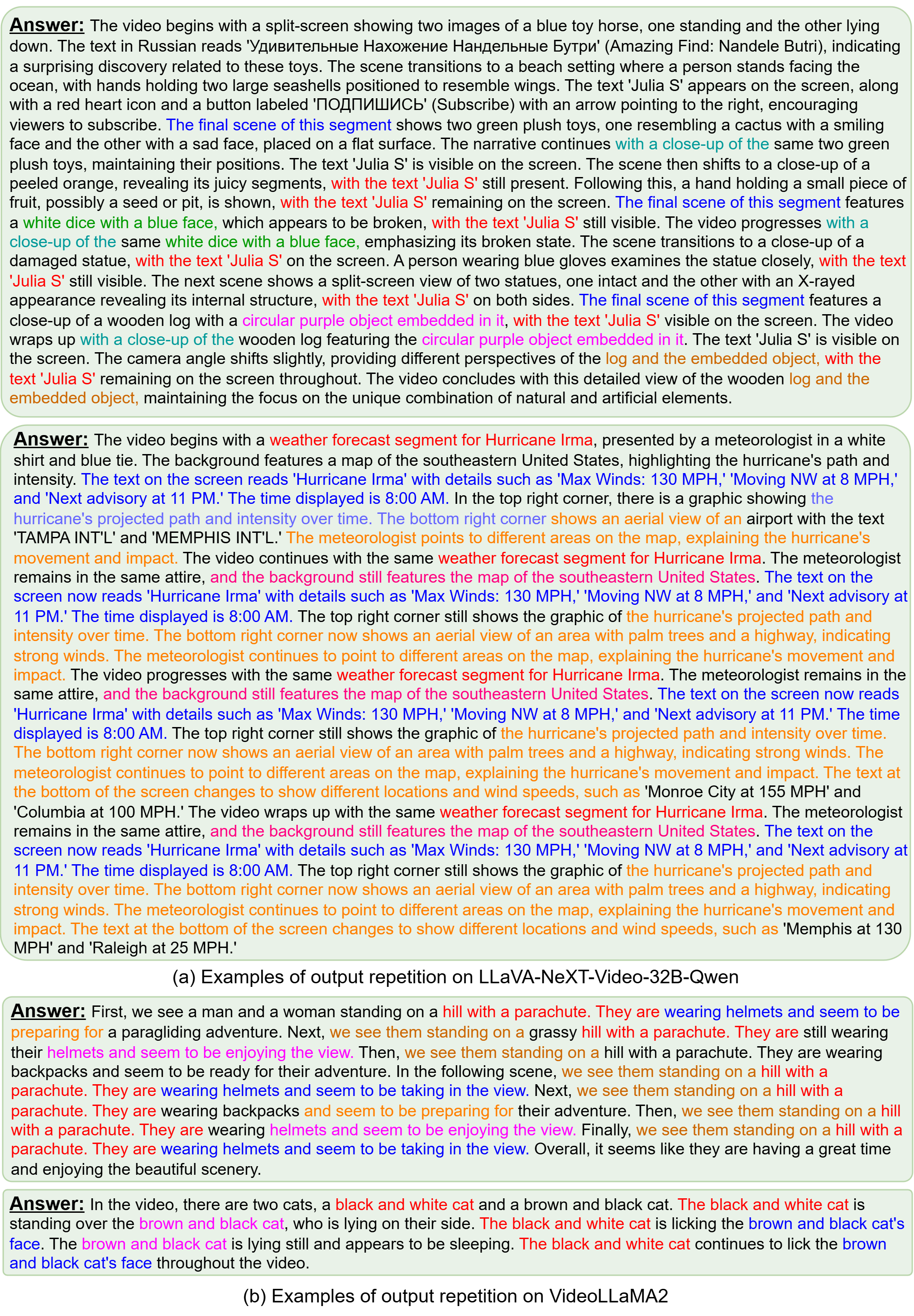}
    \caption{Examples of repetitive outputs generated by \textit{LLaVA-NeXT-Video-32B-Qwen} and \textit{VideoLLaMA2}, with repeated phrases or sentences highlighted in different colors.}
    \label{fig:examples56}
\end{figure}

\begin{figure}[t]
    \centering
    \includegraphics[width=0.9\linewidth]{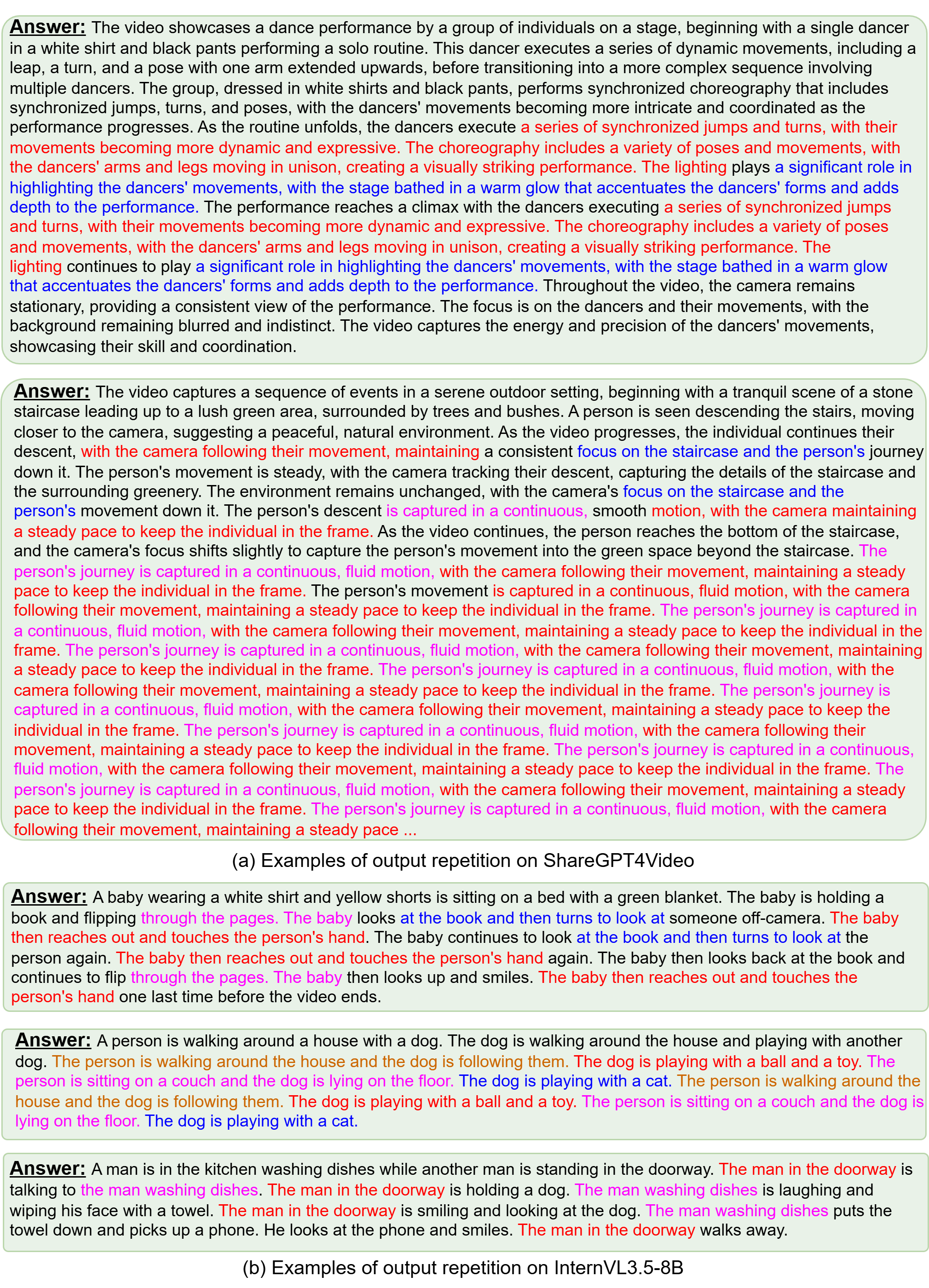}
    \caption{Examples of repetitive outputs generated by \textit{ShareGPT4Video} and \textit{InternVL3.5-8B}, with repeated phrases or sentences highlighted in different colors.}
    \label{fig:examples78}
\end{figure}

\begin{figure}[t]
    \centering
    \includegraphics[width=0.9\linewidth]{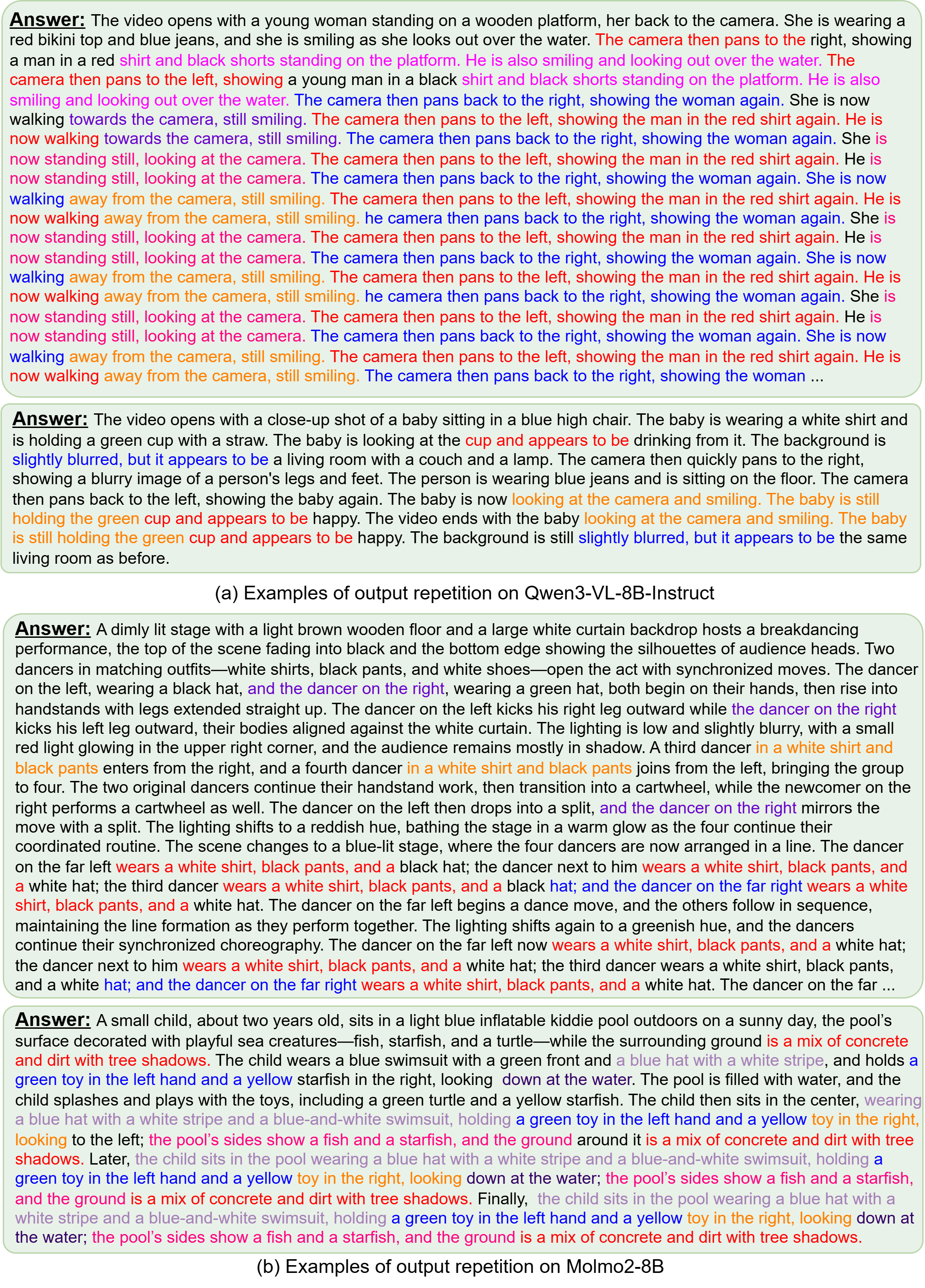}
    \caption{Examples of repetitive outputs generated by \textit{Qwen3-VL-8B-Instruct} and \textit{Molmo2-8B}, with repeated phrases or sentences highlighted in different colors.}
    \label{fig:examples910}
\end{figure}

Figures~\ref{fig:examples12},~\ref{fig:examples34},~\ref{fig:examples56},~\ref{fig:examples78}, and~\ref{fig:examples910} present representative examples of repetitive outputs from the 10 \VideoLLMs evaluated in this paper. 
As reported in the main paper, repetition is observed across all the 10 \VideoLLMs, suggesting that it is a pervasive phenomenon. 
Meanwhile, the extent of repetition varies across models, resulting in different repetition behaviors.
Models such as LLaVA-Video-7B-Qwen2-Video-Only and ShareGPT4Video often exhibit consecutive sentence-level duplication, whereas models such as LLaVA-Video-7B-Qwen2 and VideoLLaMA2 more frequently repeat shorter phrases. In addition, since LLaVA-NeXT-Video-7B and LLaVA-NeXT-Video-7B-DPO produce empty outputs at 32 frames, we further examine whether increasing the number of sampled frames correlates with abnormal generation. We find that when the sampled frame number falls within $[28,32)$, some videos still yield empty outputs, while many others exhibit highly repetitive responses that often contain grammatical errors, as shown in Figure~\ref{fig:examples34}(c). These observations indicate that these models exhibit limited decoding stability and robustness under denser frame sampling settings.

\subsection{Examples under Temporal Stressors}\label{app:transformation_repetition}
\begin{figure}[t]
    \centering
    \includegraphics[width=0.9\linewidth]{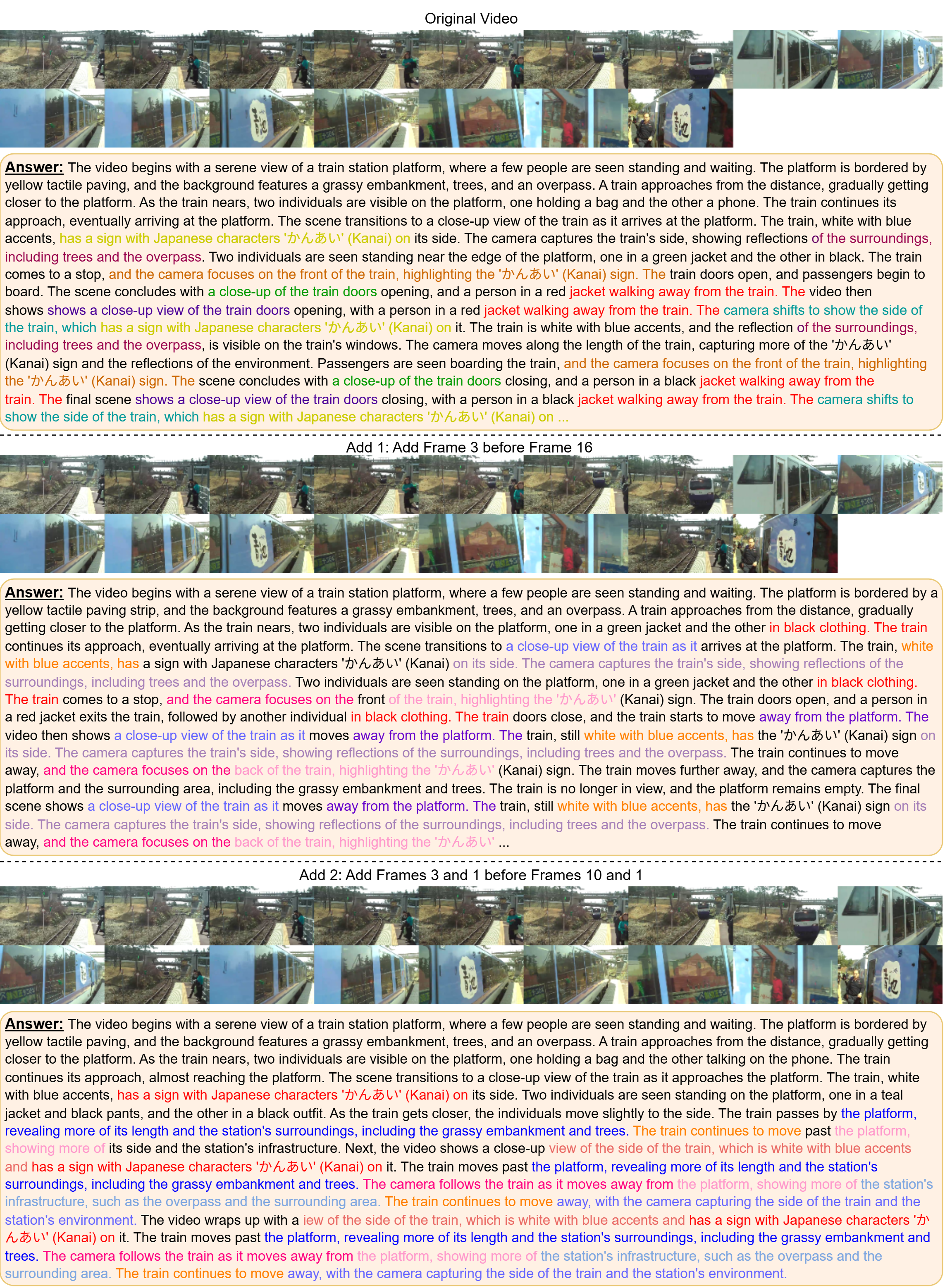}
    \caption{Example video under \textit{temporal frame insertion}. We show the sampled frames and corresponding outputs from LLaVA-Video-7B-Qwen2-Video-Only at 16 frames for the original video, as well as the videos after transformations of \textit{Add 1}  frame and \textit{Add 2} frames, with repeated phrases or sentences highlighted in different colors.}
    \label{fig:examples_add}
\end{figure}

\begin{figure}[t]
    \centering
    \includegraphics[width=0.9\linewidth]{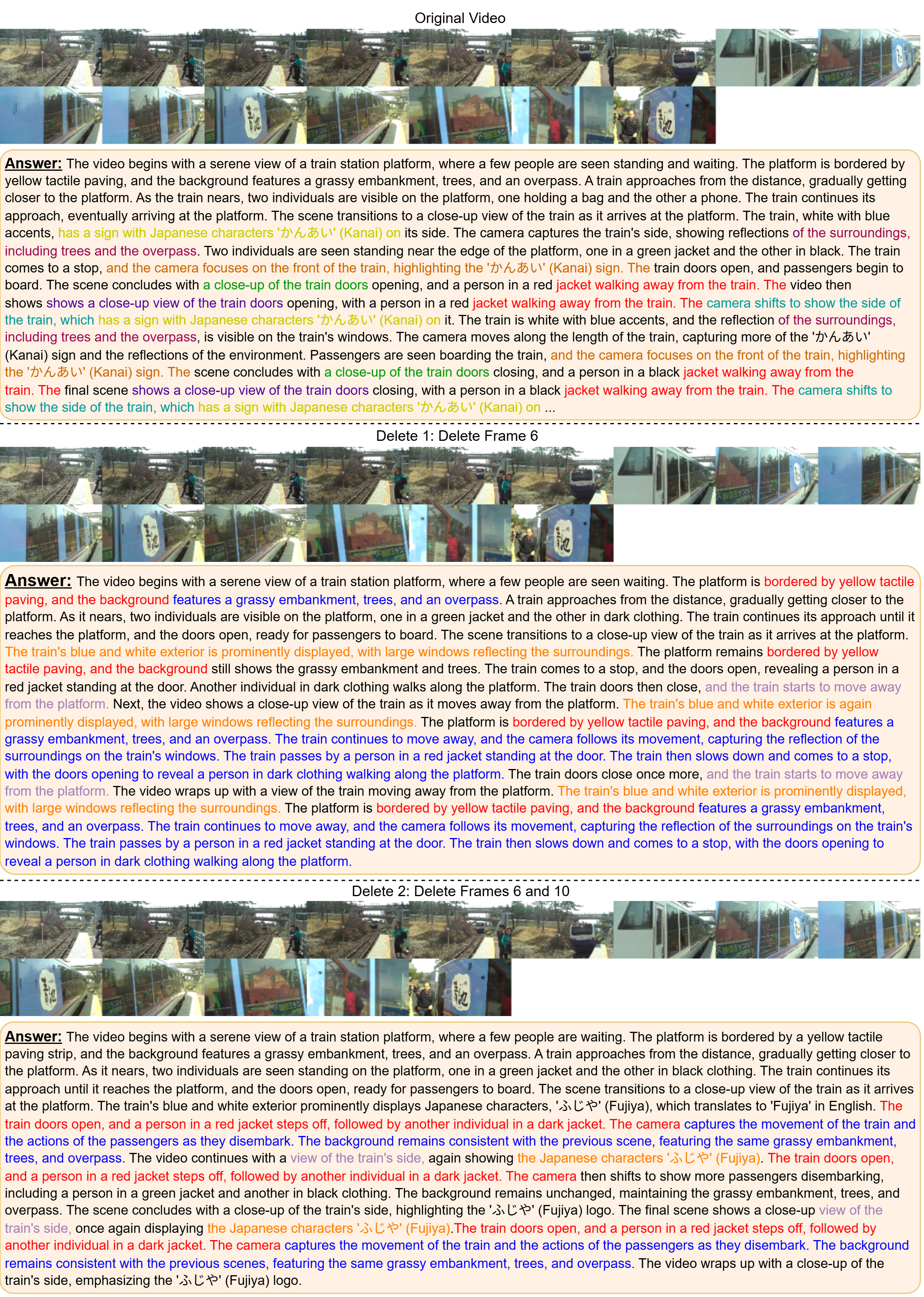}
    \caption{Example video under \textit{temporal frame deletion}. We show the sampled frames and corresponding outputs from LLaVA-Video-7B-Qwen2-Video-Only at 16 frames for the original video, as well as the videos after transformations of \textit{Delete 1} frame and \textit{Delete 2} frames, with repeated phrases or sentences highlighted in different colors.}
    \label{fig:examples_delete}
\end{figure}

\begin{figure}[t]
    \centering
    \includegraphics[width=0.9\linewidth]{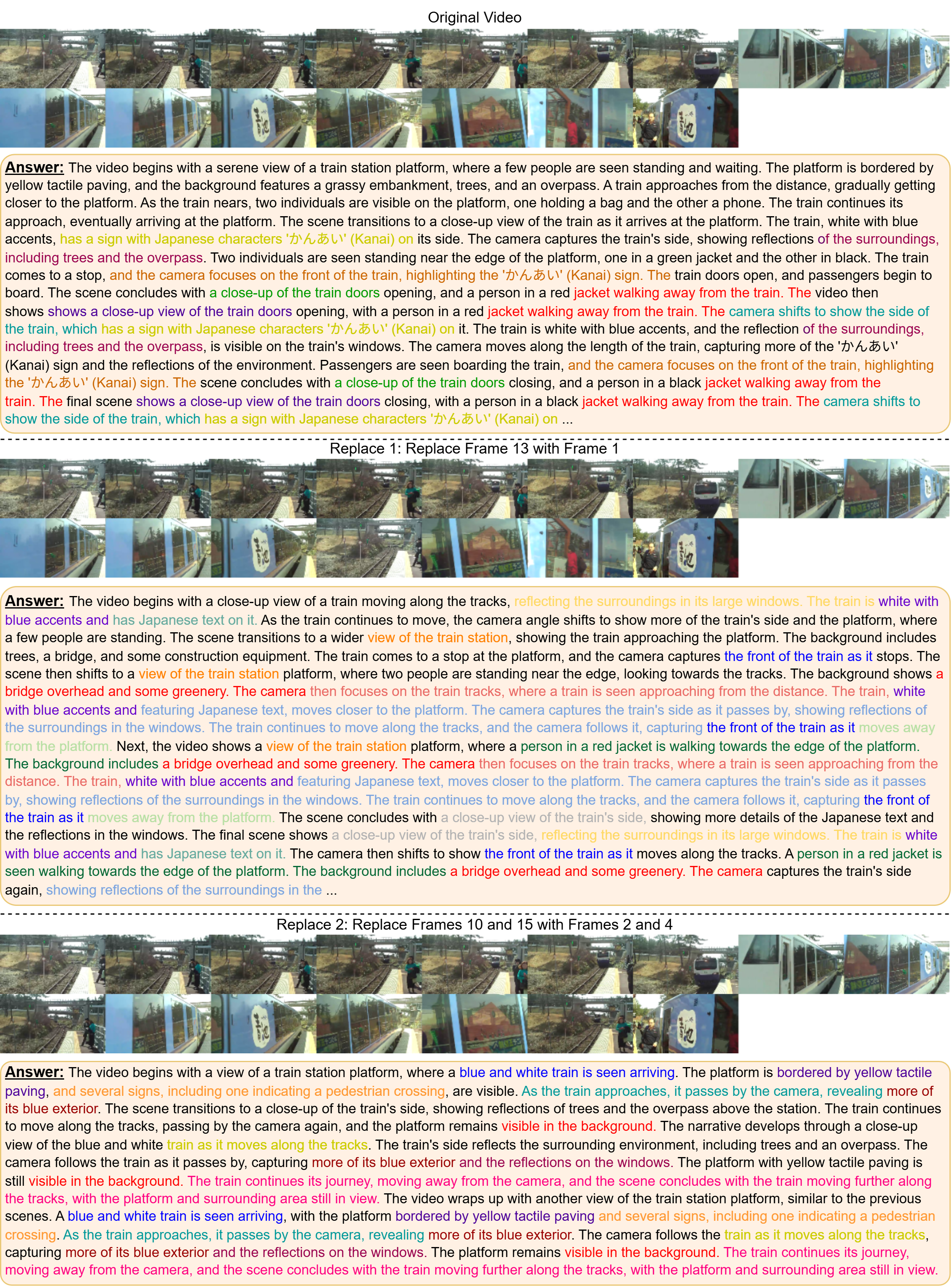}
    \caption{Example video under \textit{temporal frame replacement}. We show the sampled frames and corresponding outputs from LLaVA-Video-7B-Qwen2-Video-Only at 16 frames for the original video, as well as the videos after transformations of \textit{Replace 1} frame and \textit{Replace 2} frames, with repeated phrases or sentences highlighted in different colors.}
    \label{fig:examples_replace}
\end{figure}

\begin{figure}[t]
    \centering
    \includegraphics[width=0.9\linewidth]{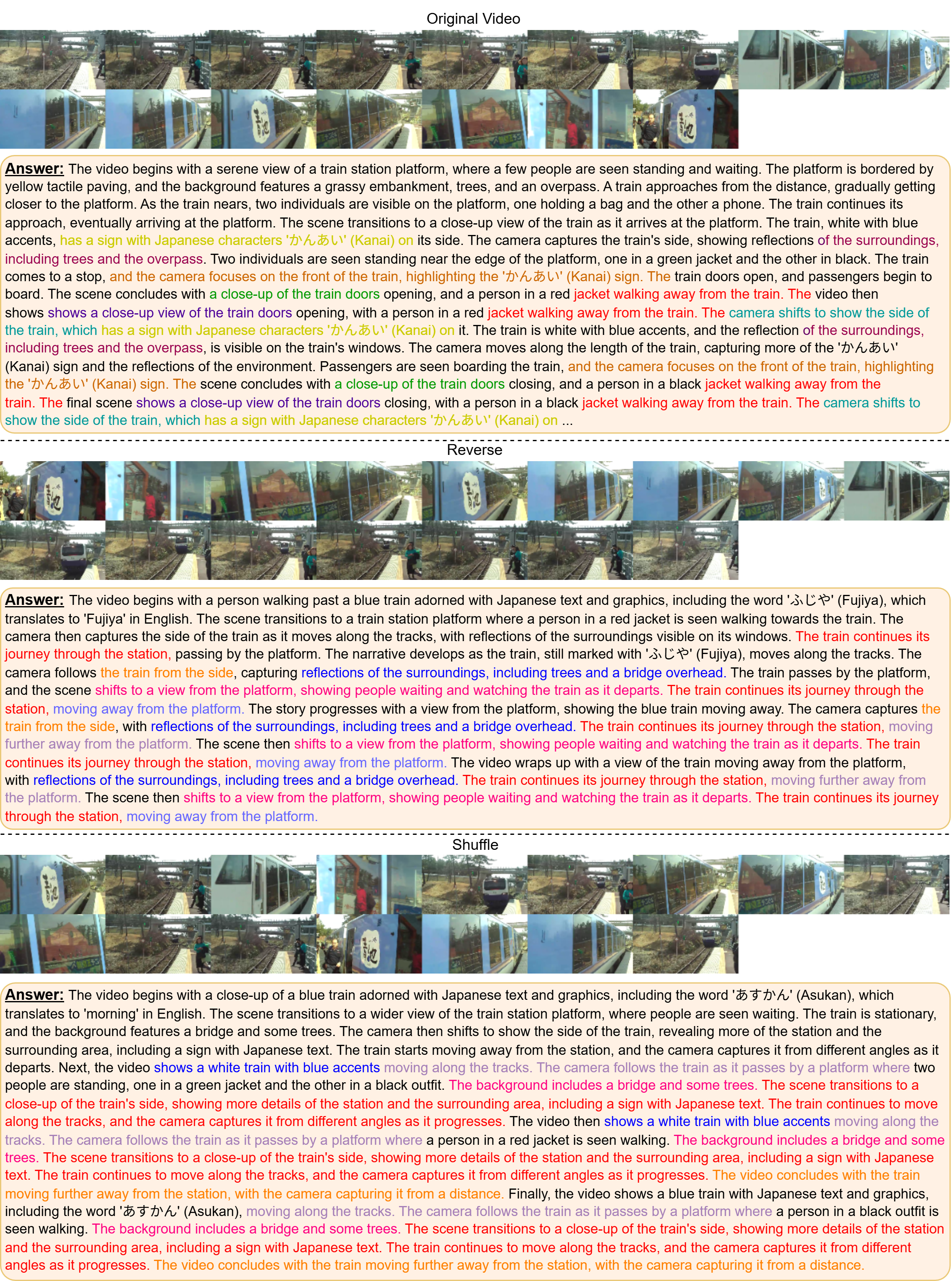}
    \caption{Example video under \textit{temporal reversal} and \textit{temporal shuffling}. We show the sampled frames and corresponding outputs from LLaVA-Video-7B-Qwen2-Video-Only at 16 frames for the original video, as well as the videos after transformations of \textit{Reverse} and \textit{Shuffle}, with repeated phrases or sentences highlighted in different colors.}
    \label{fig:examples_reverse+shuffle}
\end{figure}

Figures~\ref{fig:examples_add},~\ref{fig:examples_delete},~\ref{fig:examples_replace}, and~\ref{fig:examples_reverse+shuffle} present a representative video example, showing sampled frames and corresponding model outputs for the original video and its transformed counterparts produced by the temporal transformations in the temporal stressor library of \VideoSTF. The source video depicts a train arriving at a station. Substantial repetition is observed in the generated outputs not only for the original video but also across all transformed videos. Moreover, under several transformations, such as Add 1 and Shuffle, the repetition becomes more severe than that of the original video. This example provides supporting evidence that temporal transformations can amplify repetitive generation in \VideoLLMs even when the semantic content of the video is preserved.

\section{More Details and Results on Analysis}\label{app:analysis_details}
This section supplements \Cref{sec:analysis} with the experimental details, full results, and extended discussion behind each of its analyses. As in \Cref{sec:analysis}, the analyses that follow the settings of the main experiments are conducted on the full testbed, whereas the additional controls are conducted on an 800-video subset of it, so their absolute rates deviate slightly from those in \Cref{sec:experiments}.

\subsection{Severity Spectrum}\label{app:severity}
Repetition naturally exhibits different levels, including ordinary repetition and severe repetition (token cap reached). Ordinary repetition degrades user experience and response quality, while severe repetition leads to pathological degenerate loops that can exhaust the decoding budget. We report proportions and average token volume of benign outputs, ordinary repetition, and severe repetition in \Cref{tab:severity}. Ordinary repetition has moderate token length, while severe cases consistently reach the decoding limit (1024), showing repetition spans a spectrum, and only a subset corresponds to DoS-like behavior. Output length also matters: shorter outputs rarely become severe, but once a degenerate loop starts, length keeps growing, even when the limit is extended to 4096.
 
Our repetition metric ($n=5$, $\tau=1$) is designed to capture the \emph{onset of repetition}, rather than only extreme cases. We therefore conduct ablations with stricter thresholds (increasing the required repetition count). Even under higher thresholds, a substantial fraction of outputs remain repetitive, indicating that our conclusions do not depend on a weak definition or incidental phrasing.

\begin{table}[ht]
\vspace{-3mm}
\centering
\caption{Repetition severity statistics (\%) across models at 32 frames. Rep denotes repetition, Avg Tok is the average number of output tokens, and column $>k$ gives the share of outputs whose maximum $n$-gram count exceeds $k$.}
\label{tab:severity}
\small
\scalebox{0.8}{
\def\arraystretch{.89}
\addtolength{\tabcolsep}{-.8ex}
\begin{tabular}{lcccccccccc}
\toprule
Model & Benign & Avg Tok & Ordinary Rep & Avg Tok & Severe Rep & Avg Tok & >1 & >2 & >3 & >4 \\
\midrule
LLaVA-Video-7B-Qwen2 
& 93 & 131.54 & 7 & 188.29 & 0 & - & 7 & 1 & 0 & 0 \\

LLaVA-Video-7B-Qwen2-Video-Only
& 42 & 171.5 & 51 & 253.83 & 7 & 1024 & 58 & 27 & 14 & 8 \\

ShareGPT4Video 
& 18 & 206.06 & 65 & 356.82 & 17 & 1024 & 82 & 47 & 22 & 11 \\

VideoLLaMA2 
& 85 & 124.87 & 13 & 173.23 & 2 & 1024 & 15 & 7 & 3 & 3 \\

Molmo2-8B 
& 28 & 326.58 & 49 & 403.94 & 23 & 1024 & 72 & 32 & 13 & 9 \\
\bottomrule
\end{tabular}
}
\vspace{-3mm}
\end{table}

\begin{figure}[ht]
    \centering
    \includegraphics[width=0.99\linewidth]{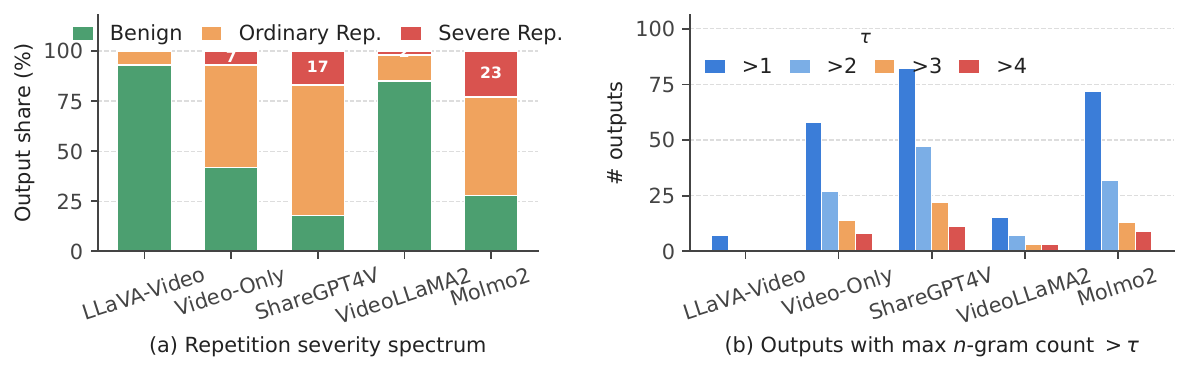}
    \caption{Repetition severity statistics across models. (a) Distribution of benign, ordinary, and severe (token-cap) outputs. (b) Number of outputs whose maximum $n$-gram count exceeds $\tau$.}
    \label{fig:severity}
    \vspace{-3mm}
\end{figure}

\subsection{Repetition and Output Quality}\label{app:quality}
On the subset, we correlate quality measures with all three \VideoSTF metrics on unperturbed videos across LLaVA-Video-7B-Qwen2, LLaVA-Video-7B-Qwen2-Video-Only, ShareGPT4Video, Qwen3-VL-8B-Instruct, and Molmo2-8B. Because the testbed is built on NExT-QA~\cite{xiao2021nextqa}, which provides question-answer pairs rather than reference captions, we adopt reference-free quality measures: the number of distinct content words, information density (distinct content words per 100 words), and CLIPScore~\cite{hessel2021clipscore}. \Cref{tab:quality_corr} reports their correlations.

\begin{table}[ht]
\vspace{-3mm}
\centering
\caption{Spearman correlation ($\rho$) between quality measures and the repetition metrics.}
\label{tab:quality_corr}
\small
\begin{tabular}{lccc}
\toprule
Quality measure & RR & RI & IE \\
\midrule
Information density & -0.76 & -0.95 & 0.92 \\
Output length (words) & 0.72 & 0.84 & -0.84 \\
Distinct content words & 0.35 & 0.40 & -0.41 \\
CLIPScore & 0.00 & 0.03 & -0.04 \\
\bottomrule
\end{tabular}
\end{table}

Repetition is strongly associated with information density, with RI at -0.95 and IE at 0.92. RR is weaker at -0.76 because it is a thresholded rate rather than a graded quantity, which motivates reporting all three \VideoSTF metrics. Output length increases sharply with repetition while the number of distinct content words changes little, indicating that longer captions carry limited additional information. CLIPScore remains nearly unchanged. This is expected because it averages over the sentences in a caption, so repeating a sentence that already matches the video has little effect on the average. \Cref{tab:quality_severity} shows the same pattern when outputs are grouped by the severity categories in \Cref{tab:severity}. Severe outputs are 2.6$\times$ longer than benign ones yet contain no more distinct content words (64.7 \vs 67.0), while information density drops by 63\%, from 46.1 to 16.9 distinct content words per 100 words (Mann-Whitney U test~\cite{mann1947test}, $p=1.1\times10^{-155}$). Thus, severe repetition uses roughly 2.6$\times$ the tokens without adding any new information to the description.

\begin{table}[ht]
\centering
\caption{Output quality by repetition severity.}
\label{tab:quality_severity}
\small
\setlength{\tabcolsep}{4pt}
\begin{tabular}{lccccc}
\toprule
\makecell[l]{Group (maximum\\$n$-gram count)} & \makecell{Share of\\outputs (\%)} & Words & CLIPScore & \makecell{Distinct\\content words} & \makecell{Information\\density} \\
\midrule
Benign (1) & 72.2 & 145 & 0.2163 & 67.0 & 46.1 \\
Ordinary (2 to 4) & 24.7 & 230 & 0.2232 & 78.8 & 34.3 \\
Severe (more than 4) & 3.1 & 382 & 0.2198 & 64.7 & 16.9 \\
\bottomrule
\end{tabular}
\vspace{-3mm}
\end{table}

Answer accuracy behaves differently. NExT-QA provides five-way multiple-choice questions on the same clips, covering causal, temporal, and descriptive aspects. At 32 frames on this subset, we split the accuracy of each model by whether its free-form description of the clip is repetitive, as reported in \Cref{tab:quality_accuracy}. We use two models spanning \Cref{tab:main_results}, namely LLaVA-Video-7B-Qwen2 and its Video-Only variant, whose RR on the full testbed at 16 frames is 5\% and 67\%. For both models, accuracy is nearly unchanged across the two groups. LLaVA-Video-7B-Qwen2 is 3.0\% more accurate on videos it describes without repetition, while the Video-Only variant is 0.2\% more accurate on videos it describes repetitively, so the direction of the difference is inconsistent. A model can therefore enter a loop during free-form description while retaining its usual multiple-choice accuracy on the same video. This suggests that repetition reflects a failure of generation stability rather than a general failure of video understanding. Multiple-choice accuracy cannot capture this behavior because the output is restricted to a single option and cannot loop. Together, the quality correlations and accuracy split show that the \VideoSTF metrics capture the stability and informativeness of descriptions, an axis not covered by accuracy-based benchmarks~\cite{ning2025video,li2024mvbench,li2025benchmark}.

\begin{table}[ht]
\vspace{-3mm}
\centering
\caption{NExT-QA multiple-choice accuracy (\%) split by whether the description repeats.}
\label{tab:quality_accuracy}
\small
\begin{tabular}{lcc}
\toprule
Model & Benign description & Repetitive description \\
\midrule
LLaVA-Video-7B-Qwen2 & 86.6 & 83.6 \\
LLaVA-Video-7B-Qwen2-Video-Only & 84.7 & 84.9 \\
\bottomrule
\end{tabular}
\vspace{-3mm}
\end{table}

\subsection{Visual-Input Controls}\label{app:controls}
The metrics of \VideoSTF quantify repetition in the output text, so whether the observed repetition is induced by video input or inherited from the base LLM must be established experimentally. In every control below, the decoder weights, prompt, and decoding configuration are fixed, while only the visual input is varied. Any resulting change in RR can therefore be attributed to the visual input. All controls are conducted on the subset, and \Cref{sec:analysis} defines each setting and summarizes the results in \Cref{tab:controls_main}, which we analyze in more detail below.

Uninformative visual input produces little repetition. Black frames yield 0\% to 11\% RR and noise frames yield 0\% to 2\%, despite supplying visual tokens without meaningful temporal content. A static video with maximal redundancy yields 0\% to 30\%, compared with 7\% to 85\% for 16 genuine frames, and no model repeats more on the static input than on genuine frames. The failure therefore requires multiple distinct real frames. Increasing frame count amplifies repetition only in susceptible models. For the three high-repetition models, LLaVA-Video-7B-Qwen2-Video-Only, ShareGPT4Video, and Molmo2-8B, RR rises from 4\%, 26\%, and 15\% at one frame to 61\%, 84\%, and 74\% at 32 frames, with most of the increase occurring by 8 frames. In contrast, the two low-repetition models, LLaVA-Video-7B-Qwen2 and Qwen3-VL-8B-Instruct, remain between 2\% and 14\% across all frame counts. \Cref{tab:main_results} covers 8, 16, 24, and 32 frames, where these curves have largely plateaued, so the frame-count stability reported in \Cref{sec:study1} remains valid for these commonly used settings.

The two LLaVA-Video variants also degenerate when asked to describe a video without visual input, reaching 56\% and 85\% RR. Inspection shows that they confabulate a frame-by-frame narration before entering a loop. This ungrounded-generation failure is distinct from the behavior measured by \VideoSTF. Supplying the video reduces RR for LLaVA-Video-7B-Qwen2 from 56\% to 7\% at 16 frames. By contrast, ShareGPT4Video, Qwen3-VL-8B-Instruct, and Molmo2-8B all have 0\% RR without video, yet two become among the most repetitive models with video input. The ranking of the five models by text-only RR is therefore largely unrelated to their ranking under video input. We also evaluate the base LLM itself because the text-only setting still uses the \VideoLLM with the visual input removed. Qwen3-8B~\cite{yang2025qwen3}, the backbone of both Qwen3-VL-8B-Instruct and Molmo2-8B, repeats on 46\% of the same prompts as a pure text LLM. Yet both derived models have 0\% RR without video and reach 12\% and 63\% under 16 video frames, showing that a shared backbone with a strong repetition tendency does not determine the behavior of the corresponding video models.

We further reduce visual information by retaining a random subset of visual tokens in their original order before decoding for LLaVA-Video-7B-Qwen2-Video-Only at 32 frames. RR decreases from 61\% with all tokens to 53\% with half and 45\% with one quarter, consistent with the frame-count result that repetition weakens as the decoder receives less visual content. Arranging the same 16 frames into a 4$\times$4 image grid yields 54\% RR, compared with 68\% when the frames are provided as a video, indicating that the image format reduces but does not eliminate repetition. The amount of visual content therefore accounts for much of the effect, while the video representation further amplifies it. We do not include image-language models as baselines because \VideoSTF targets \VideoLLMs, and repetition in image-language models is outside the scope of this study.

\begin{figure}[ht]
    \centering
    \includegraphics[width=0.7\linewidth]{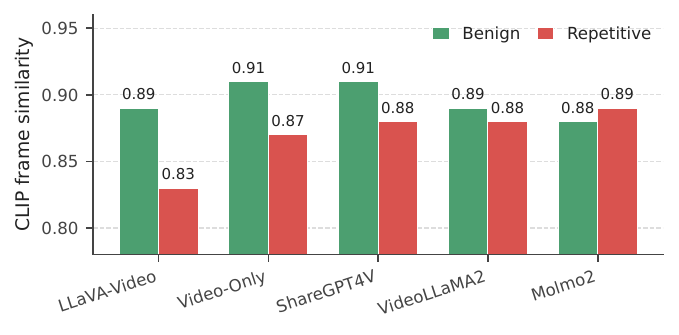}
    \caption{CLIP-based frame similarity of videos with benign \vs repetitive outputs.}
    \label{fig:clip_sim}
\end{figure}

\noindent\textbf{Input Redundancy.}
A natural hypothesis is that repetition is provoked by visually redundant inputs. As shown in \Cref{fig:clip_sim}, CLIP-based frame similarity is comparable between videos that produce benign and repetitive outputs. Because CLIP frame similarity is computed over frame pairs, it mainly captures appearance-level redundancy and is insensitive to frame order. We therefore complement it with the order-sensitive encoder-level statistics in \Cref{app:local} and the static-video control above. In that control, repeating a single frame 16 times consistently produces less repetition than using 16 genuine frames across all models, showing that maximal visual redundancy alone does not induce stronger repetition. Together with the stressor and attack results, these findings indicate that repetition remains input-dependent but is not explained simply by visual redundancy.

\noindent\textbf{Repetition on Unperturbed Videos.}\label{app:qualitative}
The mechanism analysis in \Cref{sec:analysis} focuses on stressor-induced amplification, so we separately examine why repetition occurs without any stressor and keep the two settings distinct throughout. Prevalence is measured on unperturbed videos alone, where \Cref{tab:main_results} reports RR reaching 67\%, 91\%, and 72\% for the Video-Only variant, ShareGPT4Video, and Molmo2-8B, while the temporal stress test measures sensitivity and exploitability.

On the text side, the repeated span is almost always a re-identification phrase. Most often, it names a person or object again using a color or clothing attribute, as in ``the person on the right'', ``woman in the green shirt'', or ``in a white shirt and''. A smaller but substantial group instead repeats references to a scene or spatial arrangement. Across the outputs of LLaVA-Video-7B-Qwen2, LLaVA-Video-7B-Qwen2-Video-Only, ShareGPT4Video, Qwen3-VL-8B-Instruct, and Molmo2-8B on the subset, the most frequent content words within repeated spans are white (2.9\% of content-word occurrences), shirt (2.7\%), red (2.6\%), black (2.3\%), person (2.2\%), and toy (1.6\%). No specific entity or scene consistently triggers repetition. Instead, loops form around repeated re-identification of the same entities and scenes across frames. The static-video control isolates this factor because 16 identical frames use the same frame budget and number of visual tokens as 16 genuine frames but introduce nothing new to re-identify. This reduces RR from 68\% to 30\% for the Video-Only variant. Repetition on clean videos is therefore associated with how often the description re-identifies the same content, consistent with the frame-count dependence in \Cref{tab:controls_main}.

A clip of two puppies illustrates this failure. LLaVA-Video-7B-Qwen2-Video-Only begins fluently but then cycles between the same events, with the 5-gram ``in the corner, facing each other'' appearing five times within a single description.

\begin{quote}
\small\itshape
The puppies sit in the corner, facing each other, with the bowl now empty. One puppy starts to move towards the other, and they engage in playful behavior, nuzzling and rolling around. The video continues with the puppies sitting in the corner, facing each other, and then moving towards the bowl. One puppy drinks from the bowl while the other watches. They then sit back in the corner, facing each other, and continue their playful interaction. The video concludes with the puppies sitting in the corner, facing each other, and then moving towards the bowl. One puppy starts to drink from the bowl while the other watches. They then sit back in the corner, facing each other, and continue their playful interaction.
\end{quote}

The output remains grammatical and relevant but stops adding information, repeatedly cycling between two previously described events while re-identifying the same subjects and scene. On the video side, we split the subset at the median of four per-video attributes, namely frame self-similarity, motion magnitude, shot-change count, and duration, and compare RR between the two halves. No attribute shows a consistent direction across models. Shot-change count raises RR by 24\% on ShareGPT4Video but lowers it by 16\% on Molmo2-8B, while frame self-similarity, the attribute closest to visual redundancy, yields +16\% on ShareGPT4Video and -4\% on Qwen3-VL-8B-Instruct. Neither the loop content nor these video attributes therefore identifies content that consistently induces repetition across the evaluated models.

\subsection{Local Temporal Disruption}\label{app:local}
\noindent\textbf{Temporal Disruption Versus Generic Corruption.}\label{app:corruption}
The stressors may increase repetition simply by creating unnatural inputs. To distinguish temporal disruption from generic corruption, we evaluate LLaVA-Video-7B-Qwen2 at 32 frames on the subset. We replace 2 of 32 frames with frames from a different video, or corrupt the same 2 frames using Gaussian blur, Gaussian noise, JPEG compression at quality 10, or cutout~\cite{devries2017cutout}, so that only the type of perturbation changes. \Cref{tab:corruption} reports the results. We also include a 2$\times$ speed re-sample that preserves the original frame order and introduces no temporal jump between consecutive frames.

\begin{table}[ht]
\vspace{-3mm}
\centering
\caption{Repetition rates (\%) of LLaVA-Video-7B-Qwen2 at 32 frames under frame-level changes.}
\label{tab:corruption}
\small
\begin{tabular}{lcc}
\toprule
Condition & RR & Change \\
\midrule
Original (unperturbed) & 8 & -- \\
2 frames replaced with frames from a different video & 17 & +9 \\
Gaussian blur on the same 2 frames & 10 & +2 \\
Gaussian noise on the same 2 frames & 9 & +1 \\
JPEG compression (quality 10) on the same 2 frames & 8 & 0 \\
Cutout on the same 2 frames & 8 & 0 \\
2$\times$ speed re-sample (real order, no jump) & 4 & -4 \\
\bottomrule
\end{tabular}
\end{table}

Blur, noise, JPEG compression, and cutout make the selected frames less natural but have little effect on RR. Only replacement with content from another video clearly increases repetition despite modifying the same number of frames. This pattern points to semantic discontinuity rather than generic corruption, while the lower RR under the order-preserving re-sample supports the same interpretation. Repetition on clean videos also argues against unnatural inputs alone, since \Cref{tab:controls_main} shows that repetition reaches its full level without any stressor as more frames are added.

Add and Delete modify both the frame count and the temporal structure, while \Cref{tab:controls_main} shows that frame count itself affects repetition. We therefore compare each stressor with a uniform re-sample at the same frame count on the subset, which changes the number of frames without introducing discontinuity. \Cref{tab:count_matched} reports the results. Add exceeds its count-matched control by 8\% and 6\% on the two models, indicating an additional effect from temporal disruption beyond the two extra frames. Delete is 4\% below its matched control on both models, indicating that its effect is explained by the sampling change. Cross-video replacement, which preserves the frame count, provides the clearest separation from the unperturbed condition.

\begin{table}[ht]
\vspace{-3mm}
\centering
\caption{Repetition rates (\%) of Add and Delete against count-matched uniform re-sampling.}
\label{tab:count_matched}
\small
\begin{tabular}{lcc}
\toprule
Condition & LLaVA-Video-7B-Qwen2 & LLaVA-Video-7B-Qwen2-Video-Only \\
\midrule
Original, 32 frames & 8 & 61 \\
Add 2, 34 frames & 18 & 55 \\
Uniform re-sample, 34 frames & 10 & 49 \\
Delete 2, 30 frames & 10 & 57 \\
Uniform re-sample, 30 frames & 14 & 61 \\
\bottomrule
\end{tabular}
\end{table}

\noindent\textbf{Encoder Token Statistics.}
To examine whether temporal redundancy introduced by transformations leads to repetitive outputs, we conduct additional experiments that analyze redundancy at the encoder token level after temporal transformations, rather than at the frame similarity level. Specifically, we measure three quantities:  (1) adjacent-frame token similarity, capturing local temporal redundancy; (2) global token similarity, reflecting global temporal redundancy; and (3) effective rank, measuring token diversity. We then evaluate their changes under each transformation and assess their correlation with the corresponding increase in repetition rate.

Our results show that: (1) although repetition increases under most temporal perturbations, the redundancy signals remain extremely small in magnitude; (2) global token similarity and effective rank show very weak correlation with repetition increase (Pearson corr: 0.11 and -0.076), indicating that increased global redundancy is not the driving factor;
(3) adjacent-frame token similarity slightly decreases in absolute terms, but its relative change is strongly correlated with repetition (Pearson corr: 0.81). These results suggest that stressors raise repetition by reshaping short-range temporal token interactions, which may destabilize cross-frame grounding, rather than by making inputs more redundant. Because the analysis correlates changes over eight stressors, it identifies the feature-level signal that tracks amplification and does not by itself establish a cause.

It is also worth noting that our temporal transformations, Add, Delete, and Replace, modify only one or two frames per video and do not create highly repetitive inputs. The insertion positions are random, making it very unlikely to produce consecutive identical frames. This also explains why adjacent-frame token similarity slightly decreases in absolute terms, as temporal perturbations disrupt learned motion continuity and introduce feature-level misalignment between neighboring frames, rather than making adjacent frames more similar. For Reverse and Shuffle, although they globally reorder frames, they do not introduce duplicate or repetitive content and instead break temporal coherence more drastically, so a decrease in adjacent-frame token similarity is also expected.

\begin{table}[ht]
\vspace{-3mm}
\centering
\caption{Feature-level changes and their correlation with repetition.}
\label{tab:feature_corr}
\small
\setlength{\tabcolsep}{4pt}
\begin{tabular}{lcccc}
\toprule
Trans & $\Delta$ Rep & $\Delta$ Adj Sim & $\Delta$ Glob Sim & $\Delta$ Rank \\
\midrule
Add 1 & 0.08 & -0.0010 & 8.1E-05 & -0.0036 \\
Add 2 & 0.07 & -0.0021 & -3.7E-05 & -0.00070 \\
Delete 1 & 0.08 & -0.00016 & 1.9E-05 & -0.0038 \\
Delete 2 & 0.06 & -0.00037 & 6.1E-05 & -0.0092 \\
Replace 1 & 0.05 & -0.00099 & -0.00017 & -0.0065 \\
Replace 2 & 0.05 & -0.0019 & 0.00025 & -0.045 \\
Reverse & 0.03 & -1.8E-08 & 6.0E-09 & 4.3E-07 \\
Shuffle & -0.01 & -0.018 & 3.1E-09 & -6.6E-07 \\
\midrule
Corr with $\Delta$ Rep & 1.00 & 0.81 & 0.11 & -0.076 \\
\bottomrule
\end{tabular}
\end{table}

\noindent\textbf{Decoder Attention.}\label{app:attention}
As the decoder-side counterpart to \Cref{fig:feature_corr}, we record, for each generated token, the share of attention mass assigned to visual tokens, averaged across all heads and layers. We compare repetitive and benign outputs using LLaVA-Video-7B-Qwen2-Video-Only at 32 frames on the subset. As shown in \Cref{tab:attention}, visual attention drops by roughly half from the first to the last 50 generated tokens, with nearly identical trends for repetitive and benign outputs. Measured from the onset of repetition, attention decreases from 15.4\% to 10.0\%, a smaller decline than that observed for benign outputs over the same number of tokens. The decoder therefore continues to attend to the video during repetition, indicating that the failure is not explained by a loss of visual attention. Instead, the signal most consistently associated with amplification remains the disruption of short-range temporal structure between adjacent frames on the encoder side.

\begin{table}[ht]
\vspace{-3mm}
\centering
\caption{Share of decoder attention on visual tokens (\%).}
\label{tab:attention}
\small
\begin{tabular}{lcc}
\toprule
Output & First 50 tokens & Last 50 tokens \\
\midrule
Repetitive & 19.3 & 9.6 \\
Benign & 19.0 & 9.5 \\
\bottomrule
\end{tabular}
\end{table}

\subsection{Training Data}\label{app:training_data}
\noindent\textbf{Training Data Audit.}
To examine the role of training data, we audit 120,000 captions, with 20,000 sampled from each of six public training corpora, namely LLaVA-Video-178K~\cite{zhang2024LLaVA-Video}, the ShareGPT4Video mix~\cite{chen2024sharegpt4video}, ShareGPT4V~\cite{chen2024sharegpt4v}, LLaVA-Hound~\cite{zhang2024llava-hound}, LLaVA-Instruct-150K~\cite{liu2023visual}, and PixMo-Cap~\cite{clark2026molmo2}, and apply the \VideoSTF metrics directly to the caption text. \Cref{tab:training_corpus} reports the results. We report RI rather than RR for the corpora because their mean caption lengths differ by up to 7$\times$, which confounds cross-corpus comparisons of RR. The phrase rate counts occurrences of ``continues to'' per 1,000 words. These corpus-level statistics are computed on the training captions rather than the testbed. For reference, the table also lists the corresponding models and their full-testbed results at 16 frames from \Cref{tab:main_results}.

\begin{table}[ht]
\vspace{-3mm}
\centering
\caption{Caption repetition in training corpora and the corresponding models.}
\label{tab:training_corpus}
\small
\setlength{\tabcolsep}{3.5pt}
\begin{tabular}{lccclcc}
\toprule
Training corpus & \makecell{Mean\\words} & \makecell{Caption\\RI} & \makecell{Phrase\\rate} & Model & RR (\%) & RI \\
\midrule
\multirow{2}{*}{LLaVA-Video-178K} & \multirow{2}{*}{1055} & \multirow{2}{*}{0.677} & \multirow{2}{*}{1.98} & LLaVA-Video-7B-Qwen2 & 5 & 0.33 \\
 & & & & LLaVA-Video-7B-Qwen2-Video-Only & 67 & 0.48 \\
ShareGPT4Video mix & 288 & 0.387 & 0.47 & ShareGPT4Video & 85 & 0.57 \\
ShareGPT4V & 144 & 0.370 & 0.02 & VideoLLaMA2 & 15 & 0.38 \\
LLaVA-Hound & 149 & 0.322 & 0.00 & LLaVA-NeXT-Video-7B-DPO & 8 & 0.37 \\
LLaVA-Instruct-150K & 170 & 0.317 & 0.03 & VideoLLaMA2 & 15 & 0.38 \\
PixMo-Cap & 168 & 0.300 & 0.01 & Molmo2-8B & 69 & 0.50 \\
\bottomrule
\end{tabular}
\end{table}

The looping vocabulary is also reflected in the training data. Phrases that recur in model outputs are pervasive in the training targets. Among 20,000 LLaVA-Video-178K captions, ``continues to'' appears 41,688 times, or roughly twice per caption, ``the video continues'' appears 22,299 times, and ``the scene shifts'' appears 14,823 times. Per 1,000 words, this corpus is about 4$\times$ denser in these phrases than any other corpus we examine, showing that models reproduce training patterns at the phrase level. Corpus repetitiveness alone, however, does not predict model repetitiveness. Caption RI ranks the corpora as LLaVA-Video-178K (0.677), ShareGPT4Video mix (0.387), ShareGPT4V (0.370), LLaVA-Hound (0.322), LLaVA-Instruct-150K (0.317), and PixMo-Cap (0.300), but the corresponding models do not follow this ordering. The most repetitive corpus trains the least repetitive model in \Cref{tab:main_results}, while PixMo-Cap has the lowest RI and almost no looping phrases yet trains Molmo2-8B, which reaches RR 69\% at 16 frames.

The training mixture helps explain this discrepancy. LLaVA-Video-7B-Qwen2 and its Video-Only variant share the same backbone and video corpus and differ only in whether non-video data is included in the training mixture. Their RI is 0.33 and 0.48 against a caption RI of 0.677, so removing the additional data shifts the model markedly toward the repetitive style of its video captions. Together with the DPO comparison in \Cref{tab:main_results}, where RR is 15\% to 18\% without DPO and 8\% to 11\% with it for otherwise identical models, this pattern indicates that training-stage choices influence repetition more strongly than the base LLM alone. Cleaning the caption corpus by itself is not sufficient, as the PixMo-Cap case shows, so training-level intervention remains a promising direction rather than a complete solution to output repetition.

Caption length provides an additional connection. RR flags an output whenever any 5-gram repeats, so longer text is more likely to be flagged. To control for this effect, we retain only texts that reach $W$ words, truncate them to the first $W$ words, and report the share that still contains a repeated 5-gram in \Cref{tab:training_length}. The model outputs are taken from the subset. Shuffling the words within each truncated text gives a chance baseline of 0.0\% to 0.1\% across all settings.

\begin{table}[ht]
\vspace{-3mm}
\centering
\caption{Length-matched repetition rates (\%) of training captions and model outputs.}
\label{tab:training_length}
\small
\begin{tabular}{llccc}
\toprule
Source & Name & Mean words & RR at 150 words & RR at 300 words \\
\midrule
\multirow{3}{*}{Corpus} & LLaVA-Video-178K & 1055 & 25.7 & 86.4 \\
 & ShareGPT4Video mix & 288 & 2.7 & 12.4 \\
 & ShareGPT4V & 144 & 11.2 & -- \\
\midrule
\multirow{2}{*}{Model} & LLaVA-Video-7B-Qwen2 & -- & 30.0 & -- \\
 & LLaVA-Video-7B-Qwen2-Video-Only & -- & 59.4 & 100.0 \\
\bottomrule
\end{tabular}
\end{table}

Longer captions exhibit more repetition within every corpus, and even after matching length, repetition rates differ by up to 10$\times$ across corpora. At 150 words, both models trained on LLaVA-Video-178K repeat more often than the corpus itself. The repeated content also overlaps with that corpus. Among the 5-grams repeated by each model on the subset, 38.6\% for LLaVA-Video-7B-Qwen2 and 20.7\% for the Video-Only variant appear verbatim in LLaVA-Video-178K, compared with 1.9\% to 22.8\% in five corpora that these models never saw, after matching vocabulary size. The models also show little tendency to terminate early. Their end-of-sequence probability averages 0.000 over the first 50 tokens and at the first repeated 5-gram, rising to only 0.019 over the last 50 tokens. Video-caption training therefore appears to induce a description length shaped more by annotation length than by the clip itself. A model trained on 1,055-word captions learns to generate at this scale, continues beyond the point where the clip has been described, and reuses patterns from the same corpus. This evidence remains observational, and a retraining study that varies caption length is a natural next step toward establishing the causal role of caption length.

\noindent\textbf{Training Recipes Across Models.}
Repetition appears to be shaped more strongly by video-language alignment and post-training than by the base LLM alone. This pattern is clear among Qwen-based models, where RR ranges from 8--13 for InternVL3.5-8B and Qwen3-VL-8B-Instruct to 58--72 for LLaVA-Video-7B-Qwen2-Video-Only and Molmo2-8B, suggesting that the multimodal adaptation pipeline plays a major role in repetition.

Post-training also has a clear effect. LLaVA-NeXT-Video-7B-DPO consistently achieves lower RR than its non-DPO counterpart, suggesting that alignment objectives can mitigate repetition. Similarly, LLaVA-Video-7B-Qwen2 has much lower RR than its Video-Only variant. Since the two differ only in whether non-video data is included in the training mixture, removing that data shifts the model toward the repetitive style of its video captions, as shown by the audit above.

Output length is another contributing factor. Models that generate shorter responses on benign videos (\eg LLaVA-Video-7B-Qwen2, InternVL3.5-8B) are less likely to enter repetition, while more verbose models tend to exhibit longer repetition loops.

Models trained on caption-heavy or instruction-style datasets (\eg ShareGPT4Video) also tend to generate longer outputs and consequently more repetition. We do not observe a clear qualitative difference in the form of repetition. Both native multimodal and LLM-centric models produce similar phrase-level and sentence-level repetition.

\subsection{Decoding-Time Mitigation}\label{app:decoding}
\noindent\textbf{Decoding, Filtering, and Prompting.}
To test the robustness of the output repetition, we also conduct experiments by varying decoding settings, input filtering and prompts while keeping the rest of the evaluation pipeline unchanged. \Cref{fig:decoding_robust} reports ShareGPT4Video at 32 frames on the full testbed. In \Cref{fig:decoding_robust}, T denotes temperature, p denotes top-p sampling~\cite{holtzman2020curious}, k denotes top-k sampling~\cite{fan2018hierarchical}, r denotes repetition penalty~\cite{keskar2019ctrl}, TRF denotes temporal redundancy filtering and P2, P3, and P4 denote the alternative prompts ``Summarize the video content.'', ``What is happening in this video?'', and ``Explain the sequence of events.''. The results show that repetition is largely stable across decoding choices other than the repetition penalty.

\emph{Temperature.} Increasing temperature changes the randomness of the wording, but the overall repetition rate remains similar. This suggests that stochastic decoding changes local lexical variation more than it changes the underlying tendency to fall into repetitive loops. In other words, repetition is not simply an artifact of fully deterministic decoding. Our main experiments use T=0 because deterministic decoding is important for reproducibility and fair comparison across perturbation settings. This is also the default setting for most models.

\emph{Top-p and Top-k.} The effect is limited and only partially helpful. Compared with plain T0.7 p0.9, adding top-k filtering reduces repetition to some extent, and using k50 alone reduces it further. A plausible explanation is that top-k truncation removes part of the low probability tail and can therefore suppress some unstable continuations. However, this improvement is not sufficient to eliminate repetition, and the overall trend remains the same after perturbations. Thus, these decoding heuristics may weaken the symptom, but they do not address the underlying cause.

\emph{Repetition Penalty.} This is the only decoding strategy that consistently reduces repetition more substantially. For example, the repetition rate of original videos decreases from 82 to 56 with r1.1 and 27 with r1.2, and similar reductions appear in Add, Delete, and Replace. However, stronger repetition penalties distort normal token reuse, which is often necessary for fluent and faithful video description, and we quantify the resulting quality trade-off on the subset below.

\emph{Temporal Redundancy Filtering.} We achieve this by removing highly similar or duplicate frames before feeding the video into the model. However, we observe that the repetition rate only decreases slightly under this setting, and the overall behavior remains largely unchanged. This suggests that repetition is not solely caused by explicit frame-level redundancy in the input.

\emph{Prompt Variation.} It also has only a minor effect. Although the absolute values fluctuate somewhat across P2, P3, and P4, repetition remains high overall and the relative trend across original videos and transformed videos is preserved. This indicates that the phenomenon is not highly sensitive to a particular prompt wording. Instead, it appears to be a more intrinsic behavior of current VideoLLMs when processing temporally redundant or temporally perturbed inputs.

\noindent\textbf{Decoding Settings and Output Quality.}\label{app:decoding_quality}
We re-run several decoding strategies on LLaVA-Video-7B-Qwen2-Video-Only at 32 frames on the subset, add a stronger repetition penalty of 1.3, and score each setting jointly for repetition and output quality, as reported in \Cref{tab:decoding_quality}. Following \Cref{fig:decoding_robust}, T denotes temperature, p denotes top-$p$ sampling, and r denotes the repetition penalty, and Severe is the share of outputs that reach the token cap.

\begin{table}[ht]
\vspace{-3mm}
\centering
\caption{Repetition and description quality under decoding settings.}
\label{tab:decoding_quality}
\small
\begin{tabular}{lccccc}
\toprule
Setting & RR (\%) & Severe (\%) & Words & CLIPScore & Distinct content words \\
\midrule
T0 (greedy, default) & 61 & 12 & 209 & 0.2323 & 72.3 \\
T0.7 p0.9 & 60 & 20 & 227 & 0.2295 & 77.5 \\
T0.7 p0.9 r1.1 & 50 & 8 & 212 & 0.2284 & 84.6 \\
T0.7 p0.9 r1.2 & 36 & 0 & 224 & 0.2261 & 105.3 \\
T0.7 p0.9 r1.3 & 10 & 0 & 225 & 0.2196 & 121.1 \\
\bottomrule
\end{tabular}
\end{table}

Stronger penalties suppress repetition more effectively. A mild penalty of r1.1 lowers RR from 60\% to 50\% relative to sampling without a penalty, while RR falls from 61\% under greedy decoding to 36\% at r1.2 and 10\% at r1.3. The severe token-cap tail disappears entirely at r1.2. This reduction comes at a cost to visual grounding. CLIPScore decreases monotonically with penalty strength, by 1.7\%, 2.7\%, and 5.5\% at r1.1, r1.2, and r1.3 relative to greedy decoding, indicating that stronger penalties reduce agreement with the video. Over the same range, the number of distinct content words increases by 17\% to 68\%, while output length grows by only 8\%, so the captions do not simply become shorter or less informative. The penalty reduces repetition and increases lexical diversity at the expense of grounding. The same trade-off appears on LLaVA-Video-7B-Qwen2, one of the least repetitive models, where r1.3 lowers CLIPScore by 4.5\% and increases output length by 48\%. This suggests that the cost arises from the constraint itself rather than only from suppressing repetition. On one clip, greedy decoding repeats ``in the black shirt'' 13 times in 432 words, while r1.3 produces 417 words without repetition but introduces groceries and a phone that are absent from the video.

The penalty also changes what the model generates rather than how much it tends to generate. Across these settings, the output length of the Video-Only variant never decreases and remains between 209 and 227 words, while the stopping probability is unchanged at r1.3. The tendency to repeat is also model-dependent. Under the single deterministic protocol in \Cref{tab:main_results}, the 32-frame RR across evaluated models ranges from 7\% to 82\%, and the three models sharing a Qwen3-8B backbone, InternVL3.5-8B, Qwen3-VL-8B-Instruct, and Molmo2-8B, achieve 7\%, 11\%, and 72\%.

We further evaluate standard accuracy under the same settings, with results reported in \Cref{tab:decoding_accuracy}. The NExT-QA clips are the same as above, comprising 1,566 multiple-choice and 1,384 open-ended questions, while MVBench~\cite{li2024mvbench} adds 343 questions on different videos. Both multiple-choice sets are scored using a single option letter, and the open-ended answers are compared with references averaging 2.6 words. As the penalty increases, all three benchmarks produce identical answers question by question because one-letter or two-word responses cannot repeat. When we remove the single-word instruction and then request a detailed answer, the model still responds in 3 words and no answer repeats a 5-gram, indicating that long outputs arise primarily during video description. We therefore also evaluate the descriptions with a question-answering probe, where a text-only model answers NExT-QA questions using only the generated description (Caption QA). Caption QA drops from 64.1\% without a penalty to 59.5\% at r1.3, meaning that suppressing repetition reduces the useful information carried by the description by 9\%. Standard accuracy alone would suggest that r1.3 is harmless, while the description-level metrics show otherwise. CLIPScore falls from 0.2295 to 0.2196, output length remains similar, and the number of distinct words rises by 56\%. In practice, r1.2 removes 40\% of the repetition for a 1.5\% CLIPScore cost relative to sampling without a penalty, and no setting achieves low repetition while preserving the original grounding. Repetition penalties therefore provide an effective symptom-level control with a measurable and monotonic cost in visual grounding, while \VideoSTF provides the measurements needed to characterize this trade-off.

\begin{table}[ht]
\vspace{-3mm}
\centering
\caption{Repetition, description quality, and benchmark accuracy (\%) under decoding settings.}
\label{tab:decoding_accuracy}
\small
\setlength{\tabcolsep}{4pt}
\begin{tabular}{lcccccc}
\toprule
\multirow{2}{*}{Setting} & \multirow{2}{*}{RR} & \multirow{2}{*}{CLIPScore} & \multirow{2}{*}{Caption QA} & \multicolumn{2}{c}{NExT-QA} & \multirow{2}{*}{MVBench} \\
\cmidrule(lr){5-6}
 & & & & Multiple choice & Open-ended & \\
\midrule
T0 (greedy) & 61 & 0.2323 & 63.9 & 83.3 & 16.5 & 58.9 \\
T0.7 p0.9 & 60 & 0.2295 & 64.1 & 79.8 & 14.6 & 56.0 \\
T0.7 p0.9 r1.1 & 50 & 0.2284 & 62.3 & 79.8 & 14.6 & 56.0 \\
T0.7 p0.9 r1.2 & 36 & 0.2261 & 61.2 & 79.8 & 14.6 & 56.0 \\
T0.7 p0.9 r1.3 & 10 & 0.2196 & 59.5 & 79.8 & 14.6 & 56.0 \\
\bottomrule
\end{tabular}
\end{table}

\noindent\textbf{Test-Time Re-Querying.}\label{app:requery}
The obvious test-time remedies each have limitations. Input cleaning is ineffective because \Cref{app:controls} shows that redundancy does not explain repetition. As shown above, decoding-time intervention either leaves RR largely unchanged or reduces it at the cost of visual grounding. \Cref{tab:controls_main} further shows that reducing the frame count lowers RR only at 4 frames or fewer, which discards temporal content that the model is expected to use. We therefore consider a test-time remedy that preserves the weights, decoder, and frame budget, following the broader paradigm of defenses based on detection~\cite{song2025vidtoken,chen2026webagentguard,chen-etal-2026-robustness}, correction~\cite{song2024secvid}, and perturbed-input response aggregation~\cite{robey2023smoothllm} without retraining the protected model. Specifically, RR is used as a detector, and only outputs flagged as repetitive are re-queried using a different temporal view of the same video, incurring no additional cost for benign outputs. On LLaVA-Video-7B-Qwen2-Video-Only at 32 frames on the subset, \Cref{tab:requery} reports the fraction of originally repetitive outputs that become non-repetitive under each view. Cascading all four views recovers 41\% of repetitive outputs with an average of 3.18 queries per video across the subset. Most outputs remain repetitive under every view, making this a practical mitigation rather than a complete solution, consistent with a failure rooted in the model itself rather than in a particular temporal view.

\begin{table}[ht]
\vspace{-3mm}
\centering
\caption{Recovery rate (\%) of repetitive outputs under alternative temporal views.}
\label{tab:requery}
\small
\begin{tabular}{lc}
\toprule
Alternative view & Recovery rate \\
\midrule
Shuffle & 21 \\
Reverse & 15 \\
2$\times$ speed re-sample & 15 \\
Delete 2 frames & 12 \\
Cascade over all four views & 41 \\
\bottomrule
\end{tabular}
\end{table}

Overall, the repetition penalty is the only decoding, filtering, or prompting strategy we test that clearly lowers repetition, but it trades this gain for weaker visual grounding, which leaves it an incomplete remedy. Prior work on text-only LLMs~\cite{fu2021theoretical,welleck2020neural} also shows that repetition cannot be fully resolved by decoding strategies alone. Training-level changes, such as adjusting the composition and length of video-caption supervision as examined in \Cref{app:training_data}, or temporal-aware objectives that stabilize cross-frame interactions, are therefore a promising complement. These approaches require retraining models and are beyond the scope of this benchmark, so we leave them to future work.

\subsection{Video Distributions}\label{app:video_types}
\begin{figure}[ht]
    \centering
    \includegraphics[width=0.85\linewidth]{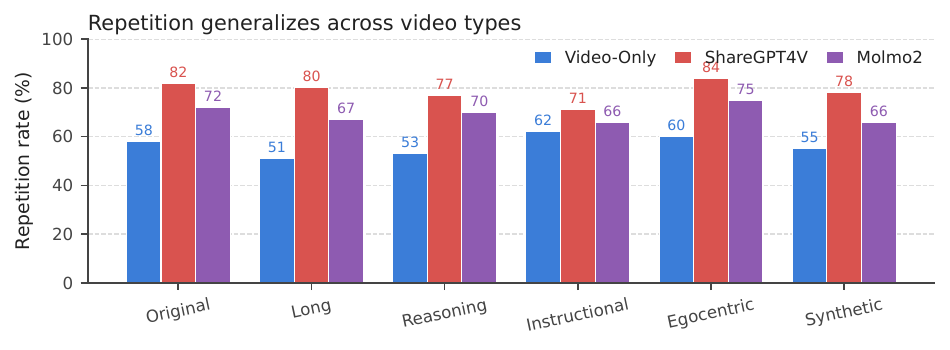}
    \caption{Repetition generalizes across video types, including long, reasoning, instructional, egocentric, and synthetic videos.}
    \label{fig:video_types}
\end{figure}

We further evaluate on longer and more complex videos by sampling from VideoMME~\cite{fu2025videomme}, Ego4D~\cite{grauman2022ego4d}, and CLEVRER~\cite{yi2019clevrer}, covering durations from 3 min to 1 h and scenarios requiring more complex temporal reasoning. As shown in \Cref{fig:video_types}, repetition consistently persists, and the three high-repetition models, LLaVA-Video-7B-Qwen2-Video-Only, ShareGPT4Video, and Molmo2-8B, exhibit comparable repetition rates on long, reasoning-heavy, instructional, egocentric, and synthetic videos. While our \VideoSTF testbed can readily incorporate these additional videos, it is important to note that the observed repetition is not specific to particular datasets or video distributions; rather, it generalizes across different settings and scales, reflecting a model-level phenomenon rather than a data-level artifact. The primary objective of this benchmark is to provide a controlled and reproducible framework for comparing models under a consistent protocol, rather than to achieve exhaustive coverage of all possible video domains and application scenarios.

\subsection{Reverse and Temporal Priors}\label{app:reverse}
Models are primarily trained on forward temporal sequences, so Add, Delete, and Replace largely preserve temporal structure, while Reverse fundamentally disrupts it. This breaks learned temporal priors and leads to qualitatively different generation behavior, making repetition less likely. This observation is consistent with prior findings (\eg Vid-SME~\cite{li2025vidsme}), which suggest that reversed temporal sequences deviate from the training distribution and are less effectively modeled by current video models. As a mitigation strategy, Reverse alone is deterministic and thus limited, and \Cref{app:requery} shows that cascading it with other temporal views recovers more repetitive outputs than any single view, although most outputs still repeat.

\section{Limitations and Broader Impacts}\label{app:limitations}
\noindent\textbf{Limitations.}
\VideoSTF is designed to measure output repetition in \VideoLLMs rather than provide a complete remedy for the failure it reveals. Among the mitigations we examine, repetition penalties reduce repetition at the cost of weaker visual grounding, while re-querying repetitive outputs with alternative temporal views recovers only a subset of them. A remedy that suppresses repetition without degrading grounding therefore remains open. The benchmark also does not propose a new \VideoLLM, and its conclusions are bounded by the 10 models and the temporal stressor families we evaluate. All of these models are open-weight, so proprietary models remain untested, and other forms of temporal change may expose further failure modes, although the consistency of our findings across model families and base LLMs suggests that the failure is widespread. Finally, the $n$-gram-based metrics are standard and widely adopted but capture surface-level repetition, so they would miss a loop that restates the same content in different words. Semantic-redundancy measures could therefore provide a complementary signal.

\noindent\textbf{Broader Impacts.}
\VideoSTF is designed to help researchers and practitioners diagnose generation failures, evaluate \VideoLLMs more comprehensively, and develop more stable video-language systems. Our attack evaluation further shows that simple temporal transformations can trigger repetitive degeneration in black-box settings, potentially degrading service quality, increasing inference cost, or affecting system availability. We include this analysis to characterize the exploitability of repetition and motivate stronger robustness rather than to encourage attacks on deployed systems. We therefore recommend responsible evaluation practices, including clear documentation of experimental procedures and disclosure of potential risks. More broadly, improved evaluation of generation stability may help identify and mitigate failures before \VideoLLMs are deployed in real-world applications.

\end{document}